\documentclass[journal]{IEEEtran}  

\IEEEoverridecommandlockouts                              

\usepackage{cite}
\usepackage[utf8]{inputenc}		
\usepackage[T1]{fontenc}
\usepackage{amsmath,amssymb,amsfonts}
\usepackage{algorithmic}
\usepackage{graphicx}
\graphicspath{{./Immagini/}}
\usepackage{textcomp}
\usepackage{comment}
\usepackage{url}
\usepackage{booktabs}
\usepackage{xcolor, soul}
\def\BibTeX{{\rm B\kern-.05em{\sc i\kern-.025em b}\kern-.08em
    T\kern-.1667em\lower.7ex\hbox{E}\kern-.125emX}}
\ifCLASSOPTIONcompsoc
 \usepackage[caption=false,font=normalsize,labelfont=sf,textfont=sf]{subfig}
\else
 \usepackage[caption=false,font=footnotesize]{subfig}
\fi
    
\begin{document}

\title{Competitors-Aware Stochastic Lap Strategy Optimisation for Race Hybrid Vehicles}

\author{Francesco Braghin$^1$, \IEEEmembership{Fellow, IEEE}, Luca Paparusso$^1$$^*$, Manuel Riani$^2$, and Fabio Ruggeri$^3$
\thanks{This work has been submitted to the IEEE for possible publication.
Copyright may be transferred without notice, after which this version may
no longer be accessible.}
\thanks{$^*$ Corresponding author. {\tt\footnotesize luca.paparusso@polimi.it}}
\thanks{$^1$ Francesco Braghin and Luca Paparusso are with the Department of Mechanical Engineering, Politecnico di Milano, Milan, Italy.}%
\thanks{$^2$ Manuel Riani is with the R$\&$D division, STMicroelectronics, Cornaredo (MI), Italy.}
\thanks{$^3$ Fabio Ruggeri is with Brembo S.p.A., Stezzano (BG), Italy.}
}

\maketitle

\begin{abstract}
World Endurance Championship (WEC) racing events are characterised by a relevant performance gap among competitors. The fastest vehicles category, consisting in hybrid vehicles, has to respect energy usage constraints set by the technical regulation. Considering absence of competitors, i.e. traffic conditions, the optimal energy usage strategy for lap time minimisation is typically computed through a constrained optimisation problem. To the best of our knowledge, the majority of state-of-the-art works neglects competitors. This leads to a mismatch with the real world, where traffic generates considerable time losses. 
To bridge this gap, we propose a new framework to offline compute optimal strategies for the powertrain energy management considering competitors. Through analysis of the available data from previous events, statistics on the sector times and overtaking probabilities are extracted to encode the competitors' behaviour. Adopting a multi-agent model, the statistics are then used to generate realistic Monte Carlo (MC) simulation of their position along the track. The simulator is then adopted to identify the optimal strategy as follows. We develop a longitudinal vehicle model for the ego-vehicle and implement an optimisation problem for lap time minimisation in absence of traffic, based on Genetic Algorithms. Solving the optimisation problem for a variety of constraints generates a set of candidate optimal strategies. Stochastic Dynamic Programming is finally implemented to choose the best strategy considering competitors, whose motion is generated by the MC simulator. Our approach, validated on data from a real stint of race, allows to significantly reduce the lap time.
\end{abstract}


\section{Introduction}
\label{sec:introduction}

\IEEEPARstart{I}{n} motorsport events, strategic decisions may have a relevant impact on the result of the competition.
Particularly, with reference to World Endurance Championship (WEC) \cite{boretti_chapter_2018}, there are four vehicle categories characterised by very different performance. This results in the formation of traffic congestion, obliging the fastest category, i.e. Le Mans Prototype 1 (LMP1), to overlap other vehicles many times during the six hours of an event. LMP1 consists in hybrid vehicles that have to respect constraints set by the technical regulation \cite{regulations} for the usage of both the electrical and thermal energy. Race engineers define how to use the energy budget in an optimal way along the track with the aim to minimise the lap time. 


In the last years, many state-of-the-art works have addressed the problem of lap strategy optimisation for race electric \cite{anselma_optimal_2021, yesil_strategy_2013, betancur_heuristic_2017, herrmann_minimum_2020, borsboom_convex_2021, broere_minimum-lap-time_2021} and hybrid \cite{duhr_time-optimal_2021, salazar_real-time_2017, salazar_time-optimal_2017} vehicles. In the majority of the cases, the formulations include very detailed and complex models of the powertrain, but do not include the presence of competitors in the driving scenario, which instead represents a fundamental aspect for what said above. Competitors are therefore neglected due to the difficulties in effectively simulating their behaviour, as well as including this information into the optimisation problem, which may become burdensome and not real-time solvable. However, this may result in overtakings being performed in points along the track where a significant deviation from the pre-computed optimal trajectory is necessary, thus determining relevant time losses.

\vspace{0.3cm}

\noindent \textbf{Statement of contributions.} To bridge this gap, we propose a computationally efficient procedure to offline identify the best strategy for the energy budget utilisation along the track, taking into account realistic traffic conditions. By best strategy we refer to the one that statistically minimises the lap time while respecting the technical regulation.

Our contribution is threefold. 
\paragraph{Statistical modelling of the competitors' behaviour} we perform a statistical analysis to practically describe the behaviour of the competitors along the track in WEC events, in terms of travel time and their mutual interactions, i.e. overtaking probability. The statistical analysis is necessary since we do not know the behavioural model of the competitors. Statistical analyses of previous events have been carried out to extract the free sector times and overtaking probability distributions. The latter refers to the probability of occurrence of overtakings between two vehicle categories along the track. This information is later employed to generate realistic simulations of the competitors' behaviour.

\paragraph{Computation of traffic-free optimal strategies}
theoretically, it would be necessary to define the best points of application of the electric motor in real time according to the actual traffic conditions, but this task is computationally too expensive to guarantee sufficiently fast cycle times. Therefore, the following solution is proposed. 

We generate a set of possible traffic-free optimal solutions, which aim to minimise the lap time while respecting the constraints imposed by the technical regulation. The set of varied solution is obtained by imposing different extra constraints on the distribution of the powertrain energy budget along the track, according to engineering expertise. The optimisations are solved using Genetic Algorithms, for their computational quickness and ease of tuning. This choice is supported by a comparison with a more classical Mixed Integer Quadratically Constrained Program (MIQCP). In both cases, the ego-vehicle is described through a longitudinal model. The set of traffic-free candidate optimal strategies will be finally tested in real traffic conditions to determine the most suitable one, as explained in the next point.

\paragraph{Optimal strategy in presence of traffic}
to identify the optimal strategy in presence of traffic, we propose an evaluation metrics based on Stochastic Dynamic Programming (SDP) \cite{SDP1,SDP2}. The dataset used by SDP is generated through the Monte Carlo (MC) \cite{Alexandrov2011} simulations, relying on a multi-agent influence/reaction model and on the previously computed statistics of the competitors.

\vspace{0.3cm}

\noindent \textbf{Sections organisation.} The remainder of this paper is organised as follows. Sec. \ref{sec: Related work} presents state-of-the-art related work. In Sec. \ref{sec:competitors_performance}, we perform the statistical analysis of the past WEC events, to compute the free sector times and overtaking probability distributions. Sec. \ref{sec:vehicle_model} details the longitudinal vehicle model that is used to solve the powertrain energy budget optimisation problem. The latter, considering traffic-free conditions, is then formulated and solved in Sec. \ref{sec:off_lap_optimization}, comparing a Mixed Integer Quadratically Constrained Program (MIQCP) formulation and Genetic Algorithms. In Sec. \ref{sec:on_simulations}, we show how to generate Monte Carlo numerical simulations of the competitors' positions along the track, employing the free sector times and the overtaking probabilities distributions, as well as a multi-agent model. Finally, in Sec. \ref{sec:results}, the previously computed traffic-free strategies are combined with the traffic-aware Monte Carlo simulations, and Stochastic Dynamic Programming is employed to evaluate the statistically best strategy in presence of traffic. Finally, conclusions are discussed in Sec. \ref{sec: conclusions}.

\section{Related work}
\label{sec: Related work}
Simulating the competitors' motion in racing events is an active field of research. A realistic simulator for circuit motorsports is proposed in \cite{heilmeier_race_2018}. It includes the effects
of tire degradation, fuel mass loss, pit stops and overtaking. However, it employs a lap-wise discretisation, which is incompatible with the goal of our work. In fact, we aim to generate an optimal lap strategy that is section-wise. A discrete-event simulation model is developed in \cite{bekker:toyota}, which is suitable for decision making to define the race strategy. The track is divided into sections. For each of them, the vehicle and environment characteristics, such as the fuel mass and the air resistance penalty, are taken into account. However, the computation of the lap time is performed using a deterministic approach, without considering the uncertainty related to the interactions between the vehicles, which is our target.

Many state-of-the-art works deal with energy management for electric and hybrid vehicles under a lower-level perspective compared to our work. Time-optimal energy management and gear shift for hybrid race cars is investigated in \cite{duhr_time-optimal_2021}. Given fuel and battery consumption targets, they implement a computationally efficient algorithm to solve the problem, mixing convex optimisation, Dynamic Programming and the Pontryagin's minimum principle. In \cite{7986999}, a similar problem is solved for real-time control of the Formula 1 power unit using a two-level Model Predictive Control scheme. Minimum-lap-time optimisation for all-wheel drive electric race cars is presented in \cite{broere_minimum-lap-time_2021}. An optimal adaptive race strategy for Formula-E cars is presented in \cite{anselma_optimal_2021}. It is based on an adaptive equivalent consumption minimisation strategy (A-ECMS) approach, and compared with a global optimal benchmark provided by Dynamic Programming. In \cite{yesil_strategy_2013}, they introduce a lap strategy optimisation method based on a Big Bang - Big Crunch approach for Solar cars in long-distance races. Finally, for Solar cars, heuristic methods are compared in \cite{betancur_heuristic_2017}. One of the implemented methods is Genetic Algorithms, which is adopted in our framework.    

The works presented here accurately represent the dynamics of the vehicle powertrain. They cannot however be employed in our framework since they neglect competitors and overtakings, which affect higher-level strategic decisions.

\section{Statistical analysis of the competitors' performance}
\label{sec:competitors_performance}
One of the key contributions of the proposed methodology for lap strategy optimisation is modelling competitors' motion along the track. Modelling and simulating the behaviour of the competitors allows in fact to design and evaluate lap strategies that can be effectively actuated in realistic racing situations. In this section, we detail how to extract a set of useful statistical indices from the publicly available fraction of data collected in previous races by WEC. The set of statistical indices is meant to synthetically describe the competitors' performance during the race, and they will be lately used to simulate their motion along the track.

Being WEC events long-lasting races, the generated amount of data provides a relevant statistical basis for our scope. However, since GPS data of the vehicles are provided by Federation Internationale de l'Automobile (FIA) to every team (for their own vehicle only) during the race but they are not publicly available, the proposed approach relies on an alternative procedure. Based on the statistical distribution of sector times data, we aim to describe the behaviour of each competitor during the race in a probabilistic fashion. The following sections illustrate the entire procedure.

\subsection{Dataset and data cleaning}
In motorsport competitions, circuits are typically divided into three main sectors. The sector times indicate the time interval spent by a car in each sector of the circuit for a specific lap. Differently from the GPS data, the sector time database is freely available\footnote{\url{http://fiawec.alkamelsystems.com/}}. An example of the database structure is reported in Table \ref{tab:data_structure}, where the columns $S_1$, $S_2$ and $S_3$ contain the three sector times. 

\begin{table*}[htb]\normalsize
	\caption{General structure of the FIA sector times database}
	\label{tab:data_structure}
	\centering
	
	\begin{tabular}{cccccccccc}
		\toprule
		\# & Lap & Stop & {$S_1$} & {$S_2$} & {$S_3$} & {Elapsed} & Class & Group & Team \\
		& & & {[s]} & {[s]} & {[s]} & {[s]} & & & \\
		\midrule
		1 	& 	1	& & 33.978 & 38.779 & 32.358 & 105.115 & LMP1 & H & Porsche \\
		1 	& 	2	& & 33.846 & 37.727 & 31.753 & 208.441 & LMP1 & H & Porsche \\
		1 	& 	3	& & 33.340 & 37.789 & 36.823 & 316.393 & LMP1 & H & Porsche \\
		\vdots & \vdots &  & \vdots & \vdots & \vdots & \vdots & \vdots & \vdots  & \vdots \\
		
		77 	& 	30	& & 40.246 & 46.643 & 40.224 & 3833.398 & LMGTE Am & \vdots & Porsche \\
		77 	& 	31	& B & 40.622 & 47.265 & 45.367 & 3966.652 & LMGTE Am & \vdots  & Porsche \\ 
		77 	& 	32	&  & 121.453 & 45.261 & 38.340 & 4171.706 & LMGTE Am & \vdots & Porsche \\
		\bottomrule
	\end{tabular}
\end{table*}

\begin{figure}[tbp]
\centering
	\subfloat[$S1$.]{\includegraphics[width=0.7\columnwidth]{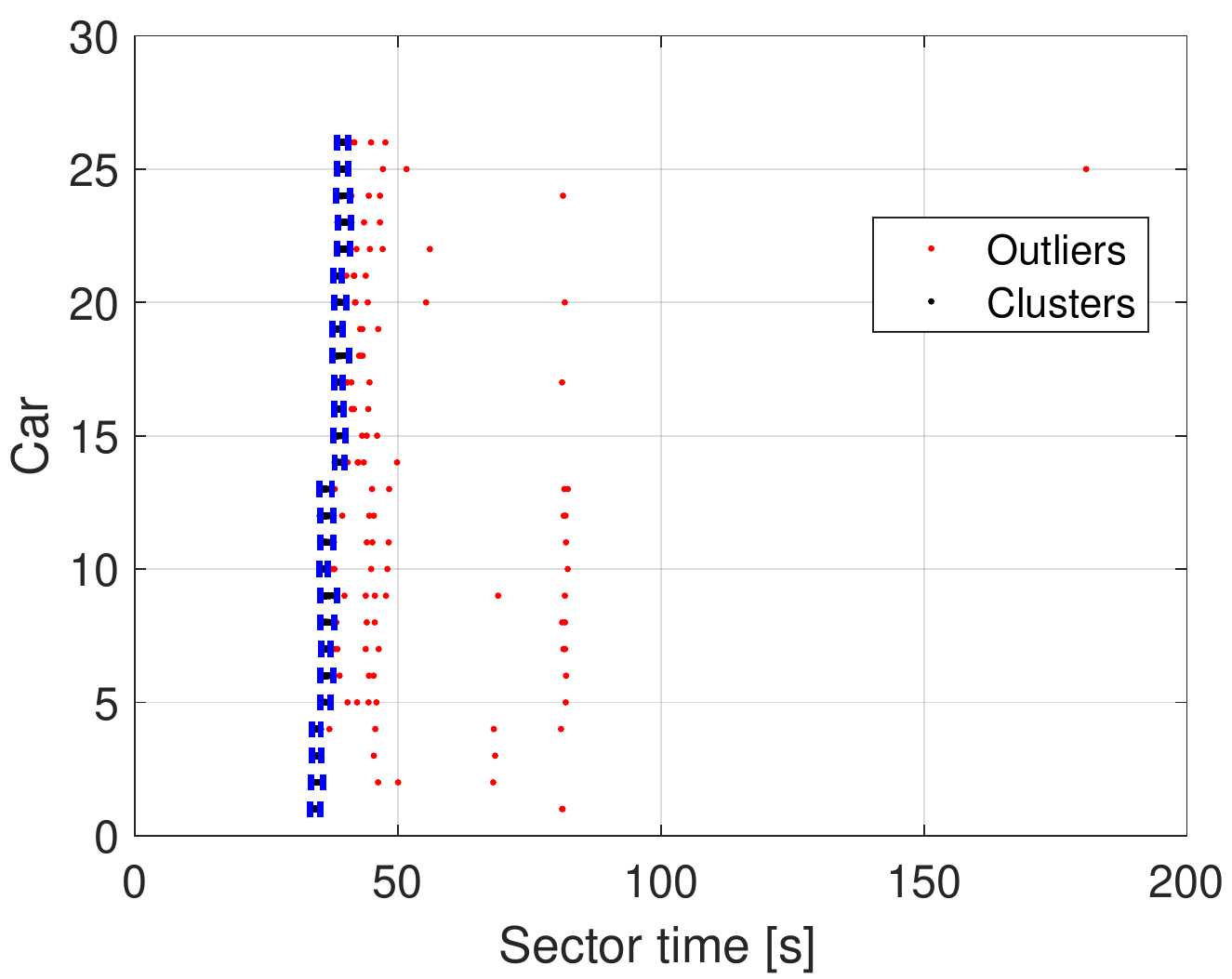}}\\
	\subfloat[$S2$.]{\includegraphics[width=0.7\columnwidth]{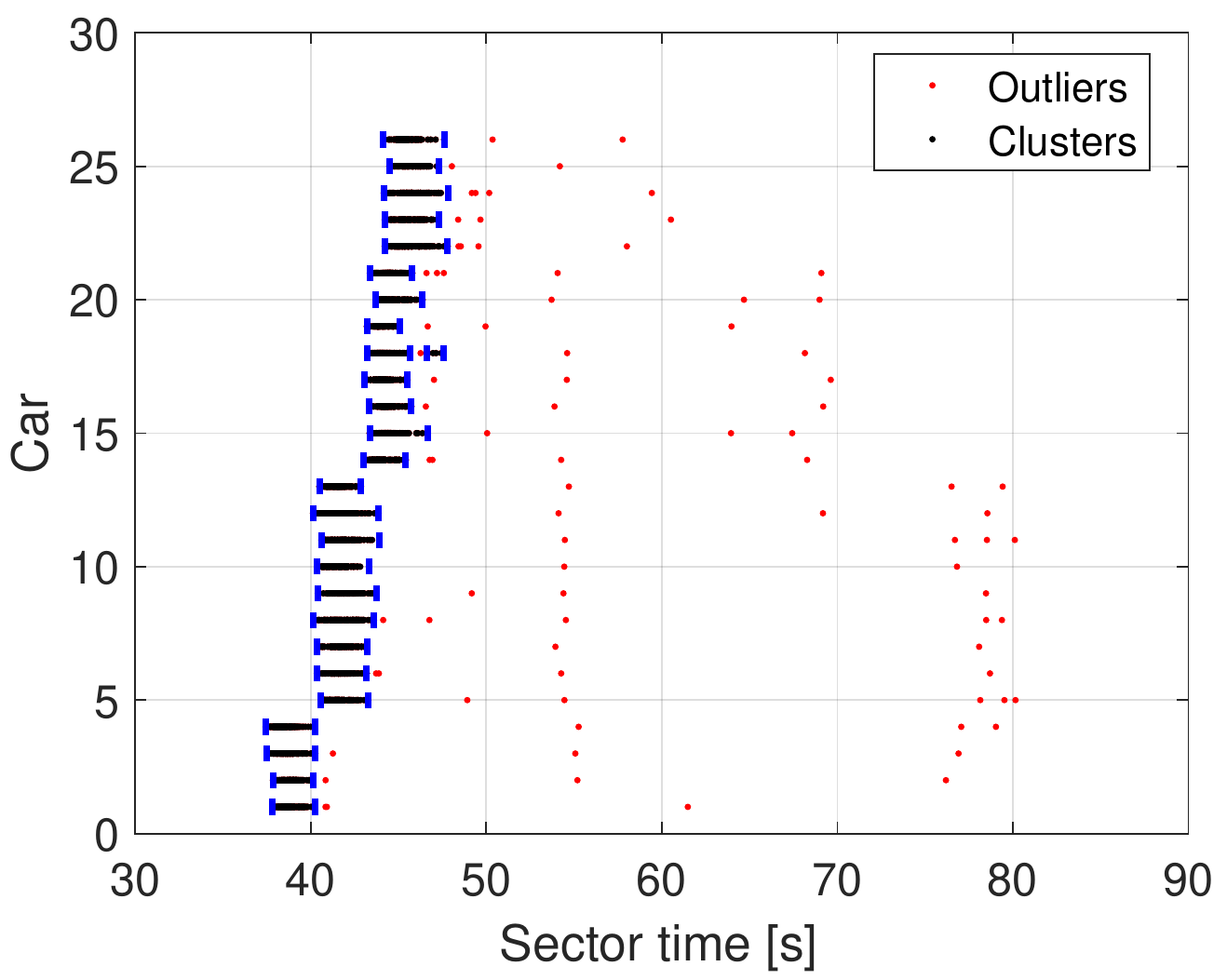}}\\
	\subfloat[$S3$.]{\includegraphics[width=0.7\columnwidth]{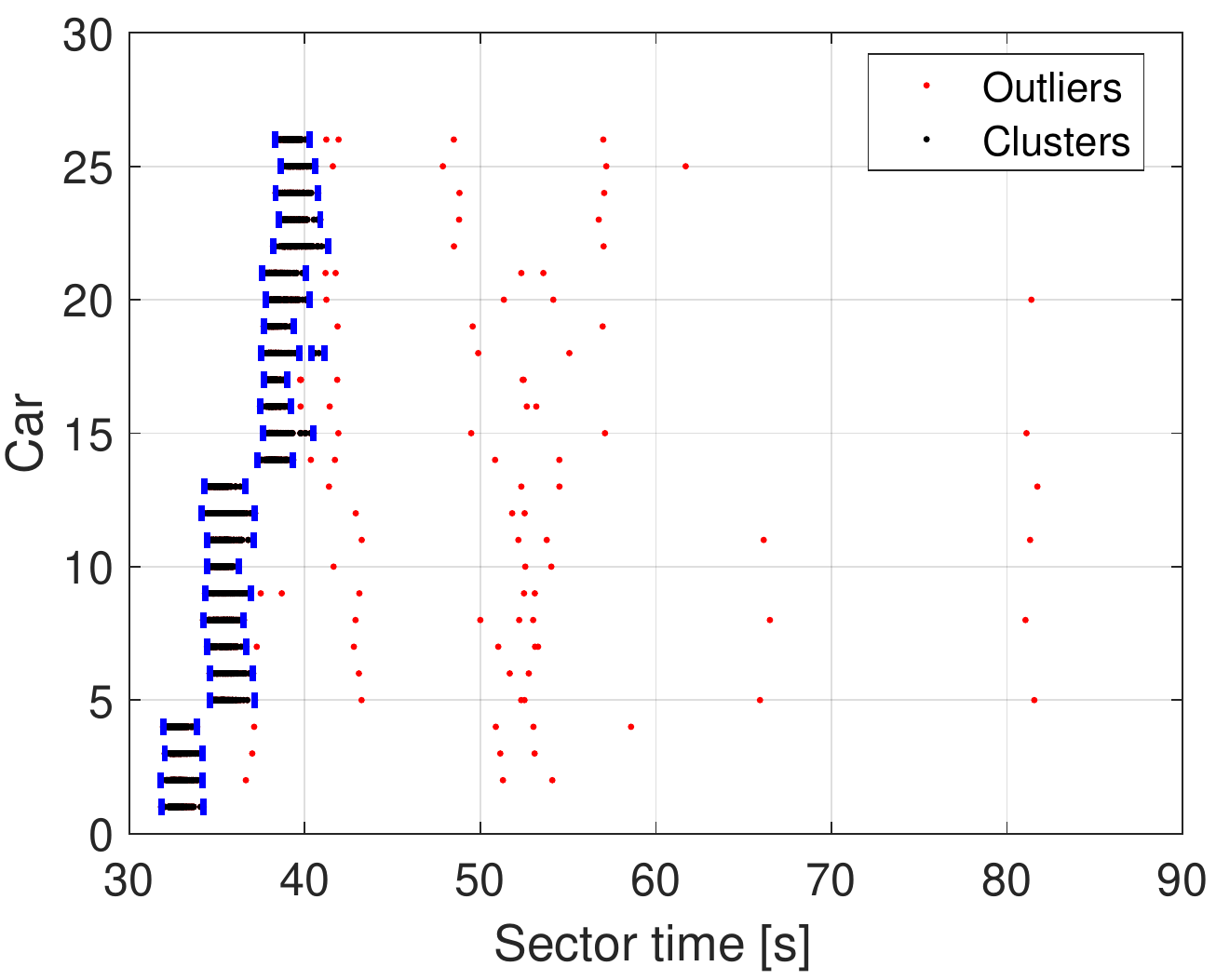}}
	\caption{Clusters subdivision of the sector times according to DBSCAN.}
	\label{fig:clusters}
\end{figure}

To eliminate spurious laps from the dataset, 
f.i. laps led by the safety car, the corresponding sector times have been clustered through the DBSCAN algorithm \cite{DBSCAN}, and only the data belonging to the cluster with the fastest sector times have been considered for successive analyses.
The clusters of fast sector times and the outliers are shown in Figure \ref{fig:clusters} for each car.

\subsection{Forecasting the competitors' speed profiles}
Resorting to the sector times, we aim to estimate the speed profile of each competitor. Then, it is straightforward to derive an estimate of the competitors position during the race, which is fundamental to statistically model their behaviour. 

\begin{figure}[htb]
	\centerline{\includegraphics[width=0.95\columnwidth]{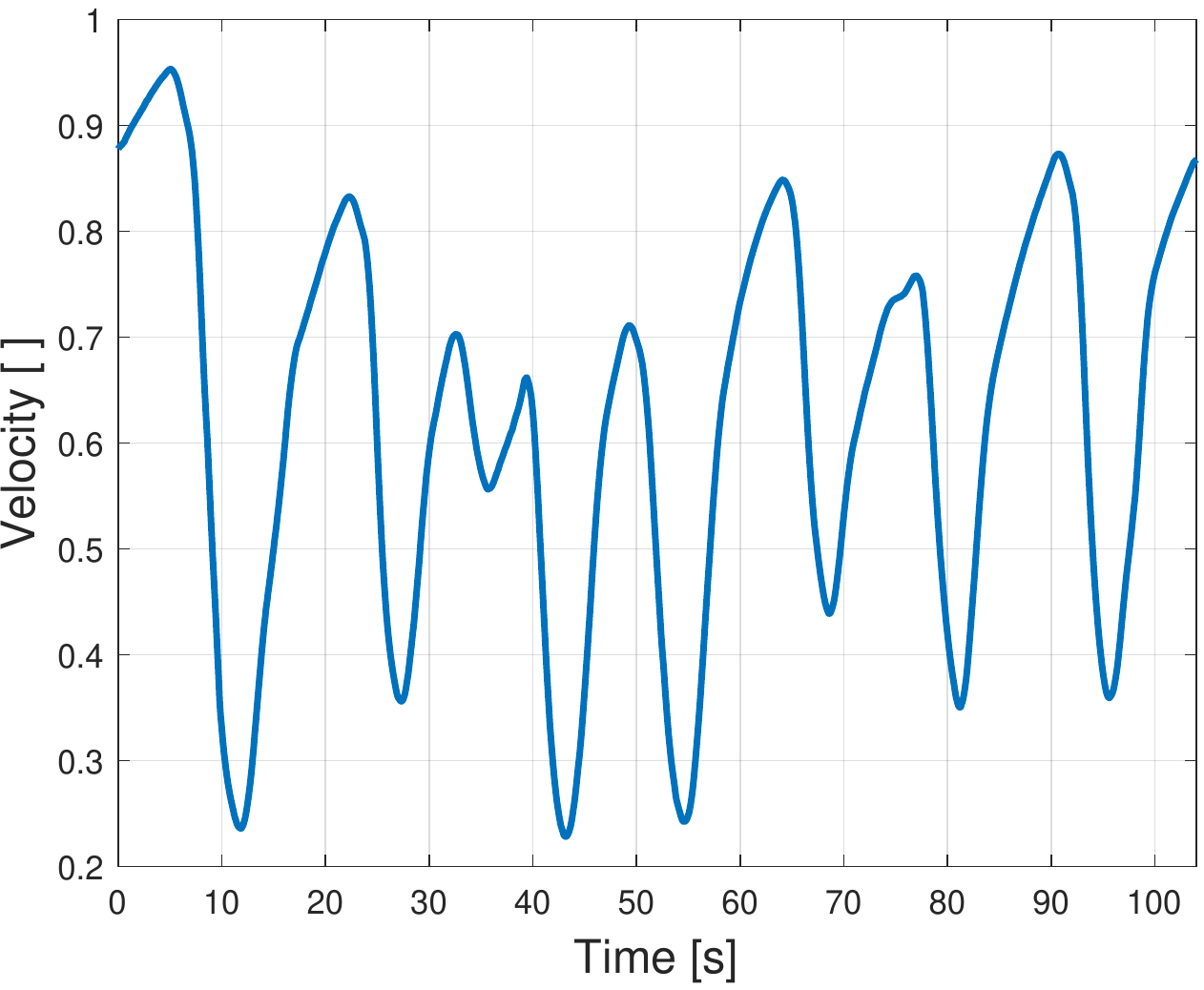}}
	\caption{Speed profile reconstructed using the linear scaling approach.}
	\label{fig:speed_profile_reconstructed}
\end{figure}

The proposed procedure to estimate the speed profiles is now described. Considering a lap performed by our vehicle in absence of traffic and without using KERS, we extract a reference profile from GPS data, as shown in Fig. \ref{fig:speed_profile_reconstructed}. We then scale it according to the measured sector times to estimate the speed profile of each competitor. The adopted linear scaling equation is
\begin{equation}
V_{c,i,j}(k) = \dfrac{V^{ref}_{i}(k) T_{c,i,j}}{T^{ref}_{i}},
\end{equation}
where
\begin{itemize}
    \item $T^{ref}_{i}$ is the $i$-th sector time performed by the car in the reference video;
    
	\item $V^{ref}_{i}(k)$ is the speed of the car in the reference video, at the frame $k$ of the $i$-th sector; 
	
	\item $T_{c,i,j}$ is the $i$-th sector time of the $j$-th lap performed by the $c$-th competitor; 
	\item $V_{c,i,j}(k)$ is the reconstructed speed of the $c$-th competitor, at the frame $k$ in the $i$-th sector of the $j$-th lap;
	
	\item $i = 1, 2, 3$ indicates the three sectors;
	
	\item $j = 1, \dots, J(c)$, with $J(c)\in\mathbb{N}_0$ being the total number of laps performed by the competitor $c$;
	
	\item $c = 1, \dots, C$, with $C\in\mathbb{N}_0$ being the total number of competitors.
	
\end{itemize}



\subsection{Statistics of the competitors' behaviour}
\begin{figure}[htb]
	\centerline{\includegraphics[width=0.95\columnwidth]{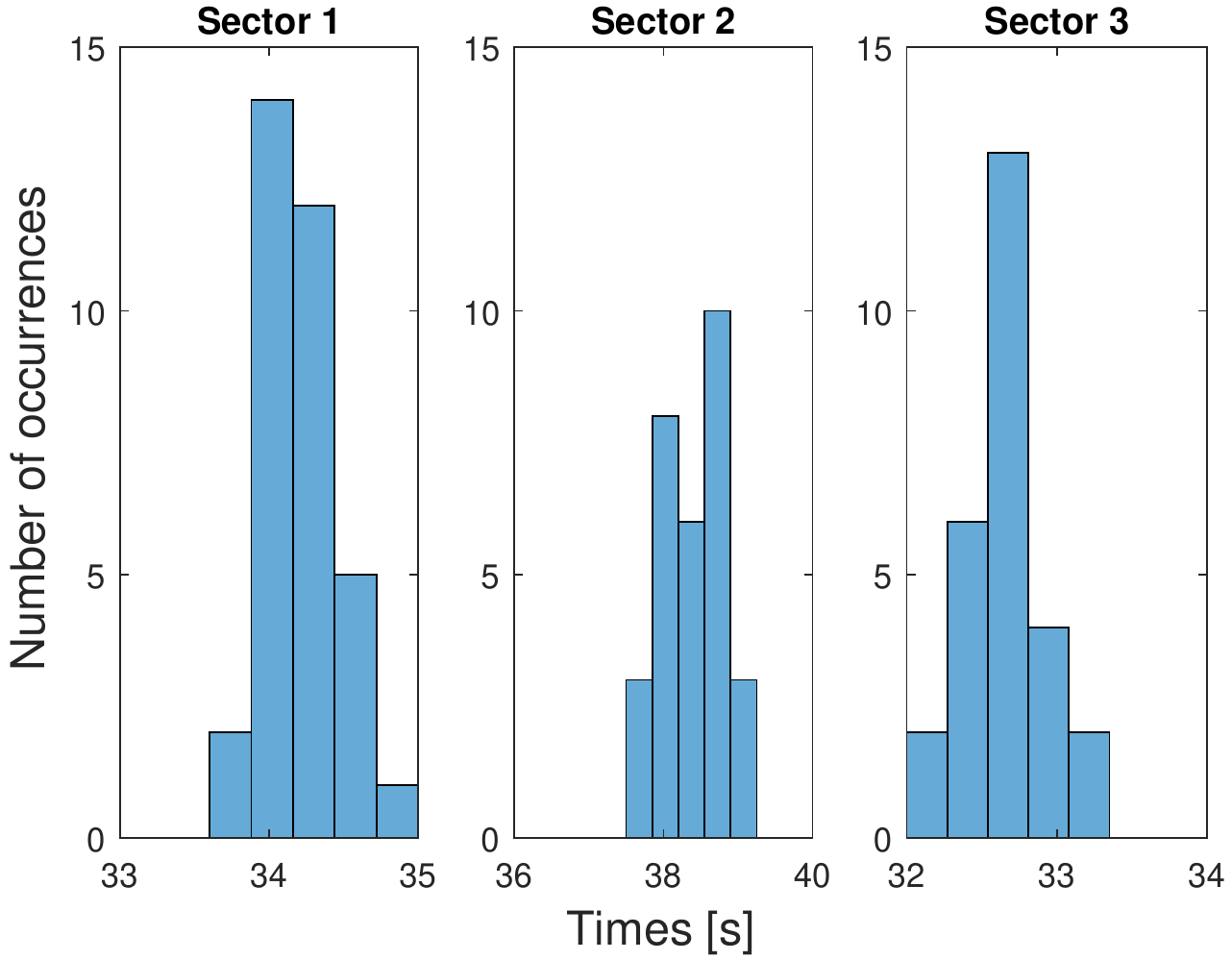}}
	\caption{Free sector times distributions for a single competitor.}
	\label{fig:FreeSectorTimes}
\end{figure}

Referring to the predicted speed profiles, the following statistics have been extracted from the cleaned dataset:
\begin{itemize}
	\item the \textit{free sector times} distributions;
	\item the \textit{overtaking probabilities}.
\end{itemize}
The free sector times are defined as the sector times performed by a competitor with the preceding vehicle far at least 100 m for the whole duration of the sector. Under these conditions, it is possible to assume that the competitor performance and behaviour have not been influenced by the other competitors. Free sector times are useful to forecast the performance of competitors in absence of interactions. In Figure \ref{fig:FreeSectorTimes}, an example of the sector times distribution is shown for a competitor. 

\begin{figure}[htb]
	\centerline{\includegraphics[width=0.9\columnwidth]{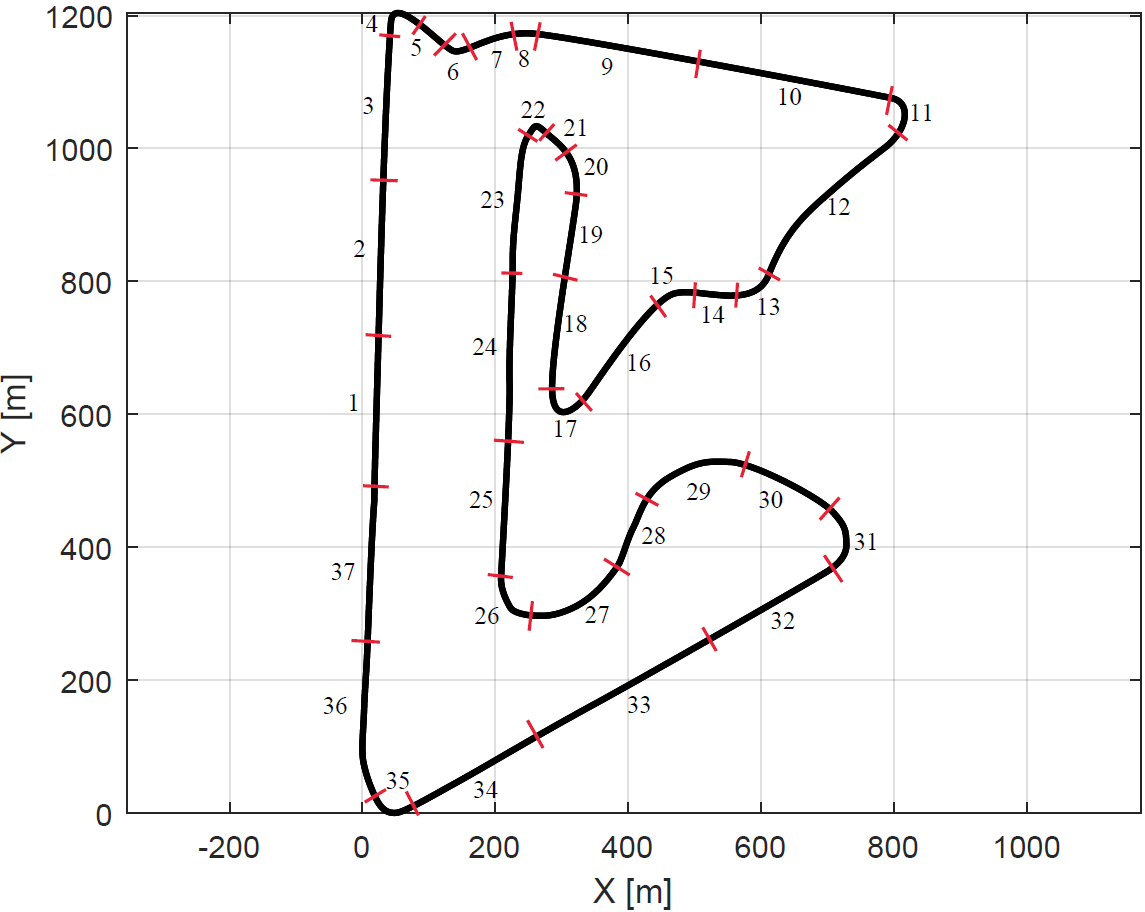}}
	\caption{Subdivision of the Bahrain circuit into sections.}
	\label{fig:sections}
\end{figure}

\begin{figure}[htb]
\centering
	\subfloat[Class LMP1.]{\includegraphics[width=0.95\columnwidth]{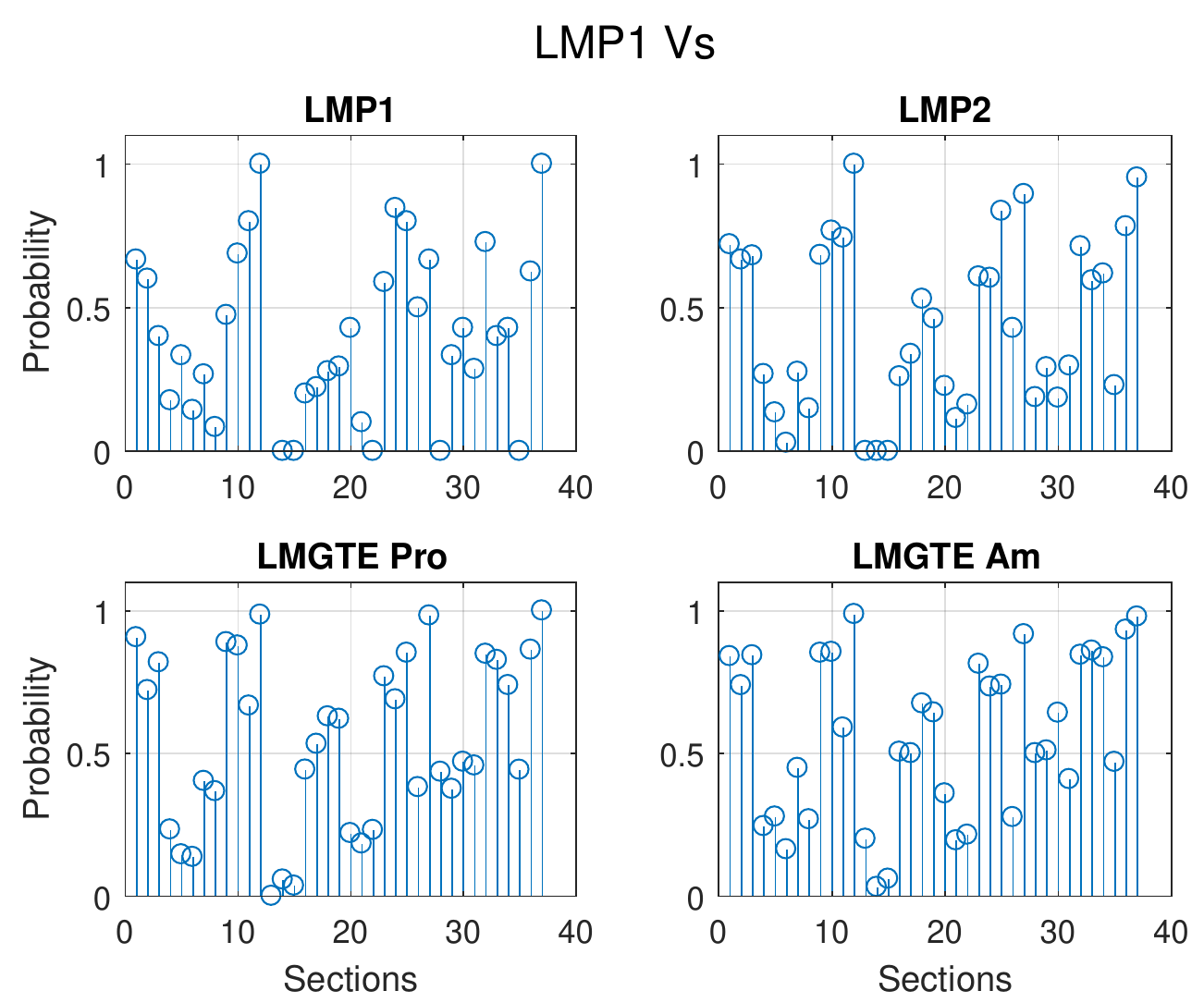}}\\
	\subfloat[Class LMP2.]{\includegraphics[width=0.95\columnwidth]{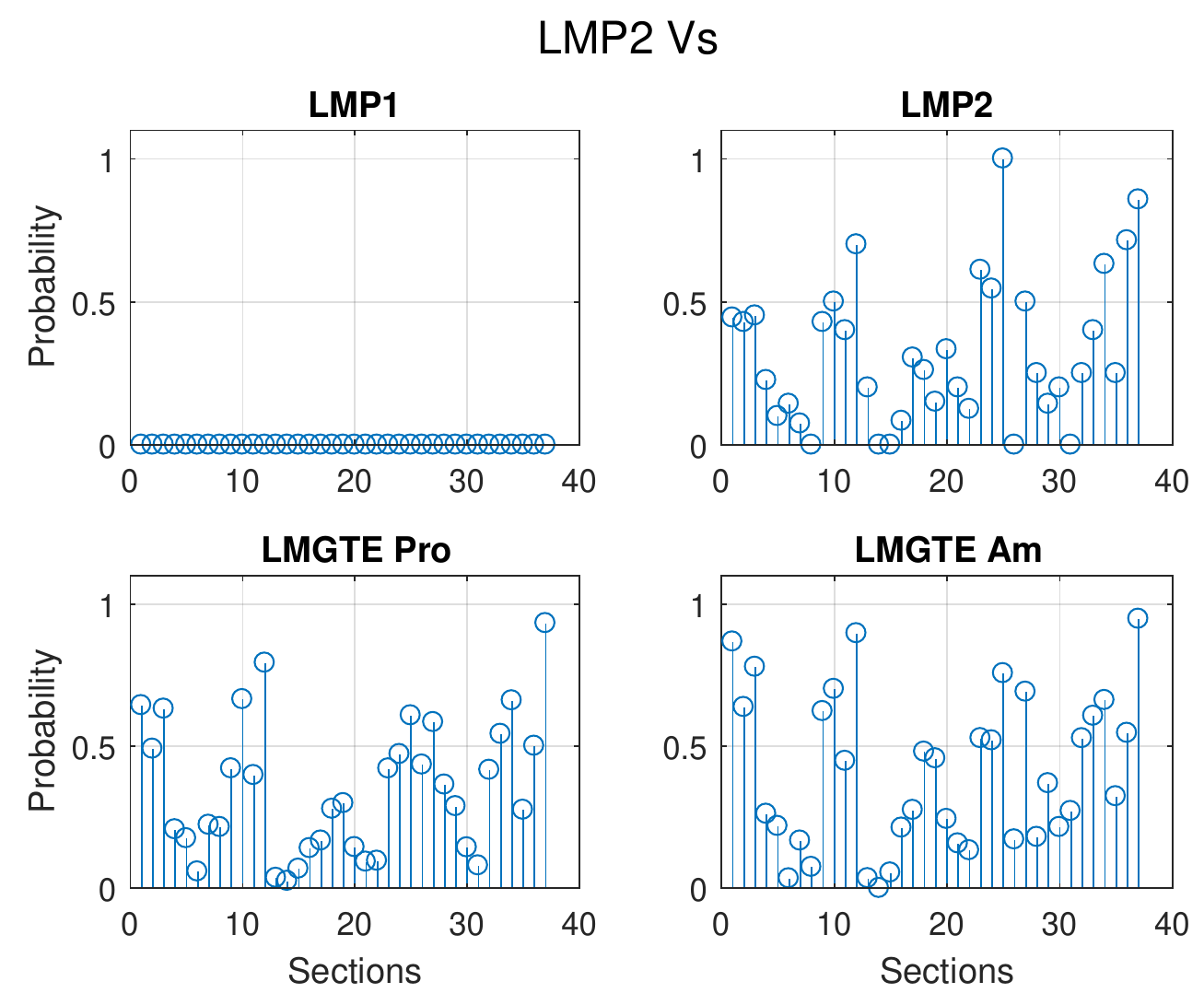}}
	\caption{Examples of overtaking probabilities distributions.}
	\label{fig:over_prob}
\end{figure}

The overtaking probabilities are modelled as functions of the section and category of the two competitors involved. These statistics are useful to practically describe the interactions between competitors along the sectors. Sections are a finer subdivision of the circuit with respect to the sectors and they allow to distinguish between straights and curves, which are typically characterised by very different probabilities of overtaking. Figure \ref{fig:sections} represents an arbitrary subdivision of the circuit of Bahrain into $37$ sections, which has been considered as the test circuit to validate the proposed approach.
Denoting with $A$ the category of the vehicle that is attempting an overtake and with $B$ the category of the vehicle that may be overtaken, the overtaking probability $\mathcal{P}(A,B,i)$ of the pair $(A,B)$ along section $i$ is computed as
\begin{equation*}
\mathcal{P}(A,B,i) =
\frac{\xi(A,B,i)}{\phi(A,B,i)},
\end{equation*}
where $\xi(A,B)$ is the number of overtakings of $A$ on $B$ in section $i$, and $\phi(A,B,i)$ is the number of times A \mbox{and} B have been in section $i$ with at most $10$ meters distance one from the other.
Let us define the following notation:
\begin{itemize}
	\item LMP1: Le Mans Prototype 1;
	\item LMP2: Le Mans Prototype 2;
	\item LMGTE Pro: Le Mans Grand Touring Endurance Professionals;
	\item LMGTE Am: Le Mans Grand Touring Endurance Amateurs.
\end{itemize}
Then, we can represent the overtaking probabilities for each pair of vehicle categories and each section. An example is provided in Figure \ref{fig:over_prob}, for the classes LMP1 and LMP2.

\section{Ego-vehicle model}
\label{sec:vehicle_model}
The statistics derived in the previous section are used to model the behaviour of the competitors, whose actual dynamics is unknown due to lack of information. Considering the ego-vehicle instead, for which we aim to design the optimal lap strategy, the necessary information to model its dynamics is supposed to be available. Therefore, we propose here a dynamic model of the ego-vehicle, which is used to study the effect of the electrical energy usage on the speed profile. In the next section, the model will be used to compute optimal lap strategies. 

\begin{figure}[htb]
	\centerline{\includegraphics[width=\columnwidth]{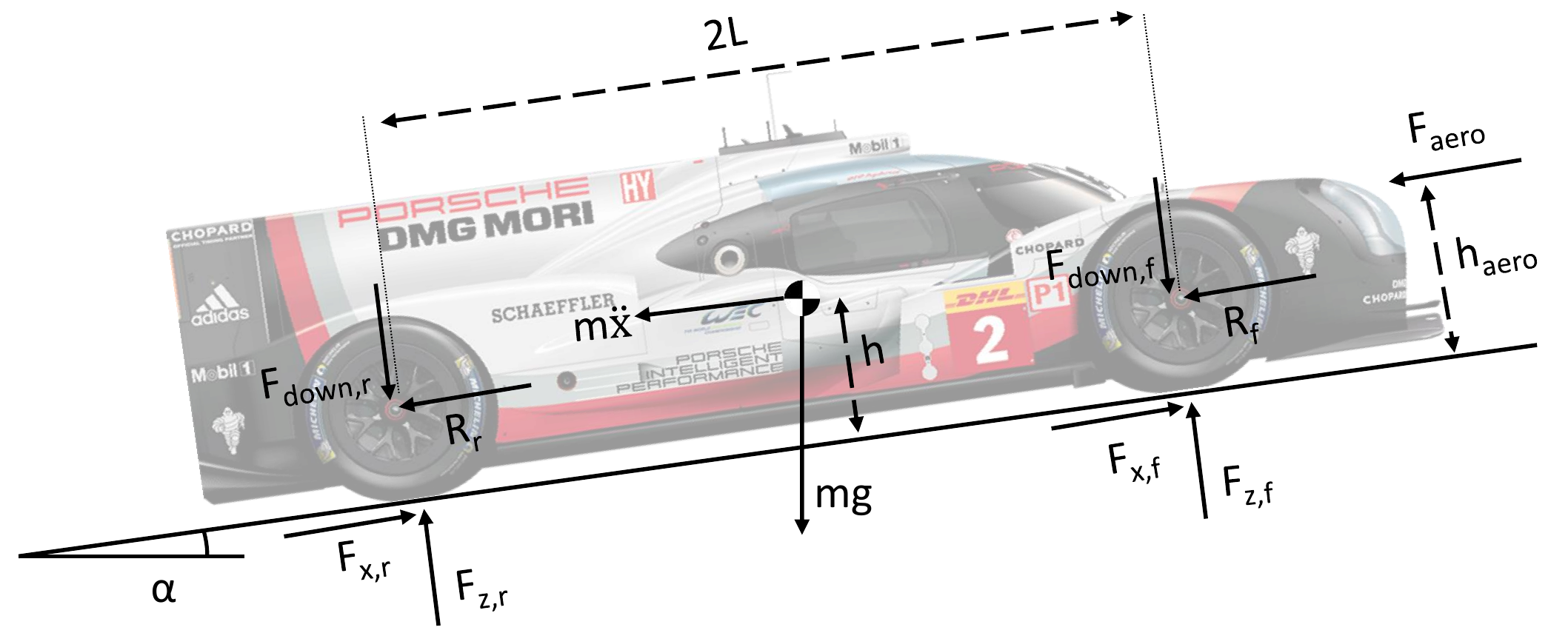}}
	\caption{Schematic representation of the ego-vehicle (LMP1).}
	\label{fig:car}
\end{figure}

The designed model is the result of a trade-off between computational complexity 
and accuracy in the description of the relevant 
dynamics. The longitudinal dynamics of the vehicle is modelled as
\begin{equation}
m \dot{v} = F_{x,f} + F_{x,r} - F_{aero} - R_f - R_r - m g \sin{\alpha},
\label{eq:EOM}
\end{equation}
where
\begin{itemize}
	\item m is the total vehicle mass, accounting also for the fuel and the driver;
	\item g is the gravity acceleration;
	\item $\alpha$ is the ground slope;
	\item $v$ is the vehicle speed;
	\item h is the height with respect to the road at which the vehicle centre of mass is located;
	\item $R_f$ is the resistive rolling force at the front wheels;
	\item $R_r$ is the resistive rolling force at the rear wheels;
	\item $F_{aero}$ is the aerodynamic resistive force;
	\item $F_{x,f}$ is the thrust applied to ground by the electric motor through the front tires; 
	\item $F_{x,r}$ is the thrust applied to ground by the combustion engine through the rear tires;
	\item $F_{down,f}$ and $F_{down,r}$ are the aerodynamic downforces, split between the front and rear tires, respectively;
	\item $F_{z,f}$ and $F_{z,r}$ are the front and rear vertical tire forces, respectively;
	\item $2L$ is the distance between the front and rear axles.
\end{itemize}
A schematic representation of the involved quantities is depicted in Figure \ref{fig:car}. 

Being the torque curves of the electric motor and of the combustion engine known, it is possible to write
\begin{equation}
F_{comb} = F_{comb} \Bigl(T_{comb}(r_{comb}), \tau_{comb}(q_{comb}) \Bigr),
\label{eq: eq comb}
\end{equation}
\begin{equation}
F_{el} = F_{el} \Bigl(T_{el}(r_{el}), \tau_{el} \Bigr).
\label{eq: eq el}
\end{equation}
With reference to \eqref{eq: eq comb} and \eqref{eq: eq el}, $F_{comb}$ represents the theoretically available thrust at the rear wheels provided by the combustion engine, $T_{comb}$ is the engine torque, $r_{comb}$ is the engine speed in rpm, $\tau_{comb}$ is the gear ratio and $q_{comb}$ is the gear selected by the driver. Similarly, $F_{el}$ represents the theoretical thrust provided by the electric motor at the front wheels, having in this case just one transmission ratio $\tau_{el}$ available. 

The amount of thrust transferred from the motors to ground depends on the maximum forces that can be generated by the tires through friction. The absolute value of the maximum tire forces $|F_{ad,f}|$ and $|F_{ad,r}|$ at the front and rear wheels, respectively, are given by
\begin{subequations}
\begin{equation}
|F_{ad,f}| = \mu |F_{z,f}|, \end{equation}
\begin{equation}
|F_{ad,r}| = \mu |F_{z,r}|, 
\end{equation}
\end{subequations}
being $\mu$ the tire-road friction coefficient. The vertical forces can be calculated through moments equilibrium as
\begin{subequations}
\begin{equation}
\begin{split}
F_{z,f} = & \,\,\,\frac{- F_{aero} \cdot h_{aero} + F_{down,f}\cdot 2L}{2L}+\\[1ex]
& + \frac{- m\dot{v}h - mgh\sin(\alpha) + mgL\cos(\alpha)}{2L},
\end{split}
\end{equation}
\begin{equation}
\begin{split}
F_{z,r} = & \,\,\,\frac{F_{aero} \cdot h_{aero} + F_{down,r}\cdot 2L}{2L} + \\[1ex]
& + \frac{m\dot{v}h + mgh\sin(\alpha) + mgL\cos(\alpha)}{2L},
\end{split}
\end{equation}
\end{subequations}
where
\begin{equation}
\begin{split}
F_{down,\cdot} &= \frac{1}{4} \rho c_z S v^2,
\end{split}
\label{eq: downforce}
\end{equation}
$\rho$ is the air density, $c_z$ is the lift coefficient and $S$ is the reference surface. The aerodynamic force is given by
\begin{equation}
F_{aero} = \frac{1}{2} \rho c_x S v^2,
\end{equation}
where $c_x$ is the drag coefficient. Finally, the resistive rolling forces are
\begin{equation}
R_{\cdot} = C_{res} \cdot F_{z,\cdot},
\end{equation}
being $C_{res}$ the rolling resistance coefficient. 

The maximum longitudinal thrust that each tire can generate through friction can be calculated by vector difference between the maximum tire force and the lateral tire force experienced during curves. The lateral tire forces $F_{y,f}$ and $F_{y,r}$ at the front and rear tires, respectively, can be computed with fair approximation considering a curve with a radius of curvature $r$ and constant vehicle speed $v$ as
\begin{equation}
m \frac{v^2}{2r} = F_{y,\cdot}.
\label{eq: lateral forces}
\end{equation}
Therefore, the absolute value of the maximum longitudinal forces $F_{t,f}$ and $F_{t,r}$ exchangeable with ground by the front and rear tires, respectively, are
\begin{equation}
|F_{t,\cdot}| = \sqrt{|F_{ad,\cdot}|^2 - |F_{y,\cdot}|^2}.
\label{eq: maximum allowed thrust forces}
\end{equation}

The driving/braking torques commanded by the vehicle powertrain depend on the usage mode. In the WEC events under analysis, there are four different modes. The vehicle central unit can decide whether to power the electric motor and combustion engine at the same time (mode 1), power the combustion engine only (mode 2), undergo sailing\footnote{Sailing is the condition for which the combustion engine is automatically powered off by the control unit to satisfy the technical constraint on the fuel usage, even if the driver applies full throttle. In this powertrain usage mode, the vehicle is decelerated by the aerodynamic forces and by the intervention of the KERS, as explained later.} (mode 3) or actuate the brakes (mode 4). Different modes imply different longitudinal tire forces, and thus different vehicle accelerations. In modes 1 and 2, the absolute value of the longitudinal tire forces can be computed as the minimum value between the thrust that the motors are capable to provide and the thrust that the tires are capable to transfer, that is
\begin{subequations}
\begin{equation}
|F_{x,f}| = \min (|F_{el}|, |F_{t,f}|), 
\end{equation}
\begin{equation}
|F_{x,r}| = \min (|F_{comb}|, |F_{t,r}|).
\end{equation}
\end{subequations}
In mode 3, the total longitudinal force is equal to the force deriving from the electric torque during sailing
\begin{equation}
|F_{x,r}| + |F_{x,f}| = |F_{sail}|.
\end{equation}
Finally, during braking the total longitudinal force is
\begin{equation}
|F_{x,r}| + |F_{x,f}| = \min(|F_{dec}|, |F_{t,f}| + |F_{t,r}|),
\label{eq: braking longitudinal force}
\end{equation}
where $F_{dec}$ is the deceleration force generated by brakes.

Considering a spatial discretisation of $2$ m, the speed profile is reconstructed using the longitudinal model and compared to a reference one.
Given the vehicle velocity and the throttle/brake commands at the current spatial discretisation point, all of the quantities are evaluated through \eqref{eq:EOM}-\eqref{eq: braking longitudinal force}, so as to obtain the vehicle acceleration $\dot{v}$, and then the velocity at the next discretisation point.
The procedure is repeated iteratively for each discretisation point.

\begin{figure}[tbp]
	\centerline{\includegraphics[width=0.8\columnwidth]{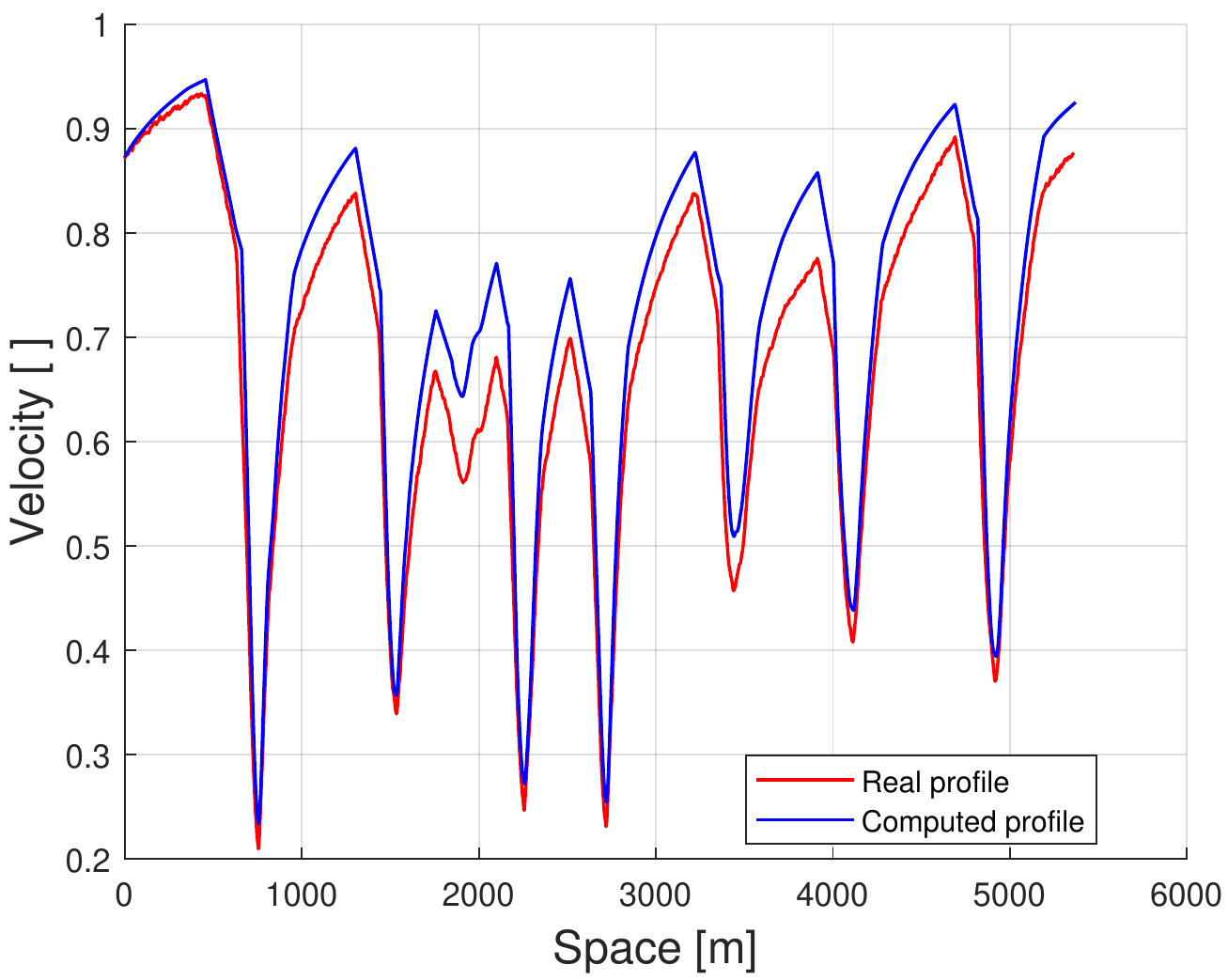}}
	\caption{Comparison between the reference speed profile and the one computed by means of the longitudinal vehicle model.}
	\label{fig:model_application1}
\end{figure}

\begin{figure}[tbp]
	\centerline{\includegraphics[width=0.8\columnwidth]{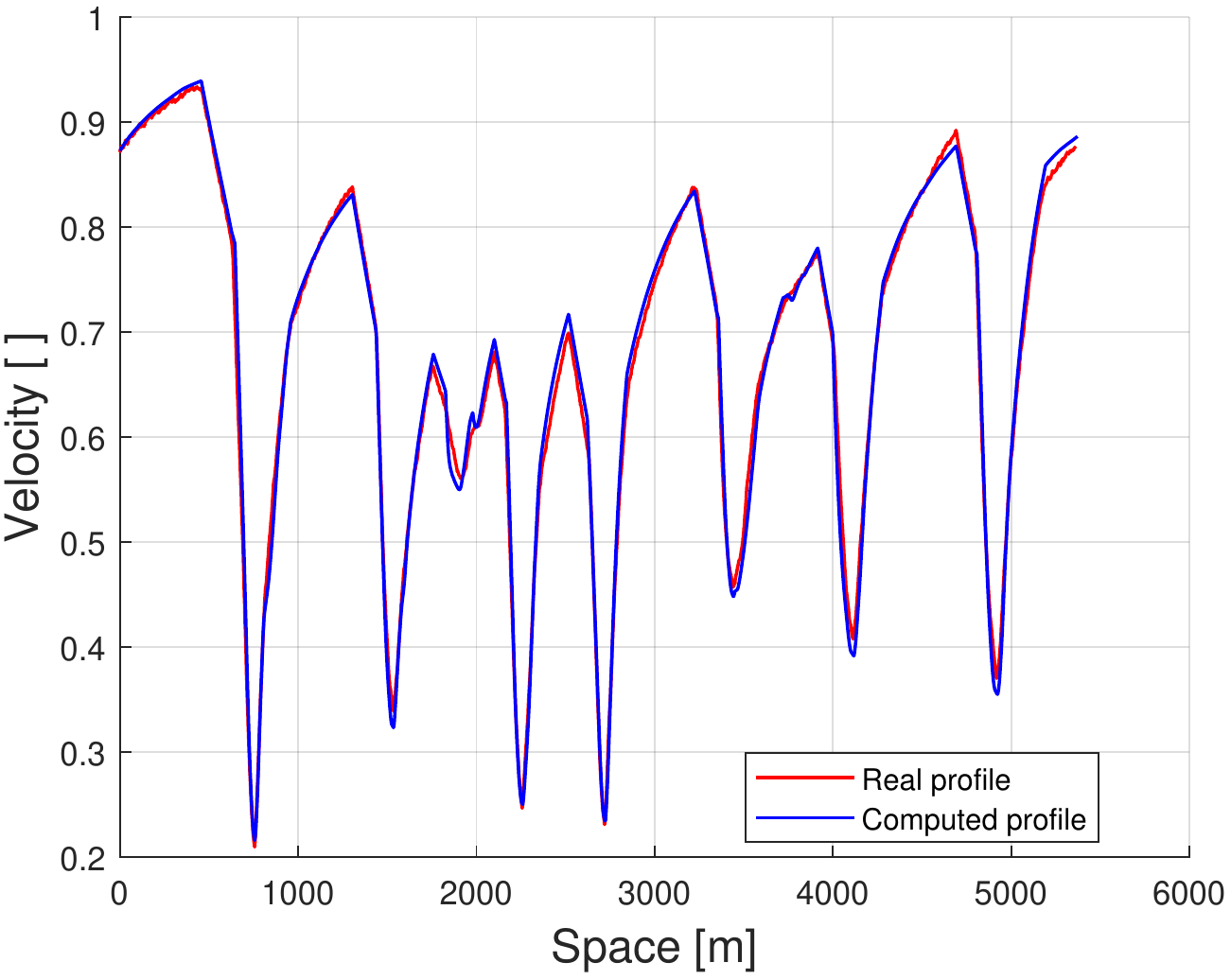}}
	\caption{Comparison between the reference speed profile and the one computed by means of the longitudinal vehicle model, after tuning the corrective coefficients.}
	\label{fig:model_application2}
\end{figure}

Hereinafter, the speed profile computed for the Bahrain circuit is compared with the real one that the reference car experienced during the real race. The results are shown in Fig. \ref{fig:model_application1}.
It is evident that the longitudinal vehicle model generates an overestimated speed profile with respect to the real one. This is due to different effects that have not been taken into account by the model. Therefore, three corrective coefficients are introduced into the model and properly tuned to better fit the real data. The corrective coefficients are:
\begin{itemize}
	\item the \textit{engine coefficient}, which scales the thrust provided by the combustion engine;
	\item the \textit{adherence coefficient}, which scales the friction coefficient $\mu$ in low-speed curves;
	\item the \textit{downforce coefficient}, which scales the lift coefficient $c_z$ in high-speed curves.
\end{itemize}
After introducing the corrective coefficients into the model, the two speed profiles result to be coherent, as shown in Figure \ref{fig:model_application2}.

\section{Traffic-free lap strategy optimisation}
\label{sec:off_lap_optimization}
Having developed and validated the ego-vehicle model, it is now possible to formulate and solve an optimisation problem for the lap time minimisation, through the management of the powertrain energy budget. We remark that, at this stage, we aim to identify the best points of activation of the electric motor in a lap, considering absence of traffic. 

Theoretically, the optimal strategy should be computed online during the real race, in order to take into account the actual traffic conditions. However, being this computation time consuming, we offline compute a set of traffic-free energy management strategies, and subsequently evaluate the best one in simulated traffic conditions (see Sec. \ref{sec:results}).

The optimisation problem was initially solved using a MIQCP formulation. Although the method can provide optimal solutions, it has many shortcomings. First of all, it is time consuming, even for the resolution of a single optimisation problem. Therefore, generating a set of offline strategies would have been intractable. Moreover, numerical issues can degrade the performance of the solver, e.g. the presence of dense columns in the resulting matrices. The solver detects and eliminates as many dense columns as possible before optimising, but this may cause numerical instability. Finally, the problem could be ill-conditioned, which may lead to inconsistent results.

These issues may be attenuated by tuning the solver parameters, which is a difficult and time-consuming operation. To cope with this, we propose an alternative method to solve the optimisation problem, based on Genetic Algorithms (GA), which provides suboptimal solutions whereas it does not suffer from the aforementioned problems. The MIQCP approach is presented just to provide a baseline for assessing the suitability of the GA. Then, the optimisation is solved multiple times with GA, varying the constraints on the electric motor usage, to generate the set of candidate lap strategies.

The basic optimisation problem, written according to the spatial discretisation, is given by
\begin{subequations}
\begin{align}
\min_{\mathbf{u}(s)} \quad & t_{lap}, \label{eq: minimization 1}\\ 
\text{subject to} \quad  & \mbox{\eqref{eq:EOM}-\eqref{eq: braking longitudinal force}},\\ 
& E_{el,used} \leq E_{el,used}^{max}, \label{eq: minimization 3}\\
& p \leq p^{max}, \label{eq: minimization 4}\\
& E_{el,rec} \geq E_{el,rec}^{min}, \label{eq: minimization 5}
\end{align}
\label{eq:minimization}
\end{subequations}
where
\begin{itemize}
	\item $t_{lap}$ is the lap time;
    \item $\mathbf{u}(s)$ is the optimisation variable, and represents the powertrain usage mode at each spatial discretisation point along the track through a one-hot encoding vector of four elements;
	\item $E_{el,used}$ is the consumed electrical energy in a lap;
	\item $E_{el,used}^{max}$ is the maximum allowed electrical energy consumption in a lap;
	\item $p$ represents the kilograms of fuel consumption in a lap;
	\item $p^{max}$ represents the maximum allowed kilograms of fuel consumption in a lap;
	\item $E_{el,rec}$ is the amount of recovered electrical energy in a lap;
	\item $E_{el,rec}^{min}$ is the minimum allowed amount of recovered electrical energy in a lap.
\end{itemize}
 
The constraints \eqref{eq: minimization 3} and \eqref{eq: minimization 4} are set by the WEC technical regulation. Referring to the Bahrain International Circuit, the corresponding limits are $E_{el,used}^{max} = 4924 \, \frac{\text{kJ}}{\text{lap}}$ and $
p^{max} ~=~ 1.381 \, \frac{\text{kg}}{\text{lap}}$. 
\vspace*{0.2cm}
Instead, the constraint \eqref{eq: minimization 5} is necessary to keep the state of charge of the battery greater than or equal to a constant value at the end of each lap. Thus, the minimum amount of recovered electrical energy must be equal to the amount of consumed electrical energy, that is
\begin{equation}
E_{el,rec}^{min} = E_{el,used}.
\end{equation}
Since the electrical energy can be recovered through both the Heat Energy Recovery System (HERS), $E_{el,rec-HERS}$, and the Kinetic Energy Recovery System (KERS), $E_{el,rec-KERS}$, we can write 
\begin{equation}
E_{el,rec} = E_{el,rec-HERS} + \text{E}_{el, rec-KERS},
\end{equation}
where the amount recovered through the HERS is track dependent and fixed for each lap.

In the next sections, the terms in \eqref{eq:minimization} will be linked to the longitudinal vehicle dynamics. Lap time optimisation based on a MIQCP formulation is detailed in Sec. \ref{sec: Mathematical Programming solver}, whereas the GA approach is presented in Sec. \ref{sec: Genetic Algorithm}. 

\subsection{MIQCP solver}
\label{sec: Mathematical Programming solver}


We decide to divide the track into $N \in \mathbb{N}$ subportions with spatial discretisation $\Delta s=5$ m. The cost function and constraints are then reformulated as functions of the optimisation variables.
\subsubsection{Cost function}
We first link the vehicle acceleration to the lap time, whose minimisation is the objective of \eqref{eq:minimization}. The kinetic energy $E_{kin}$ of the vehicle satisfies  
\begin{equation}
m \ddot{x}(s) = \frac{d E_{kin}}{ds} (s),
\end{equation}
where $s$ refers to the generic curvilinear abscissa. Identifying with $k=0,1,...,N$ the discretisation points along the curvilinear abscissa, i.e. for the generic variable $\phi$ it holds that $\phi(k):=\phi(k \Delta s)$, the above equation can be discretised using the forward Euler method, obtaining
\begin{equation}
E_{kin}(k+1) = E_{kin}(k) + m\, \Delta s\, \ddot{x}(k),
\label{eq: Ekin difference}
\end{equation}
where
\begin{equation}
E_{kin}(k) := \frac{1}{2} m v^2(k).
\label{Kinetic_Energy}
\end{equation}
Computing the vehicle speed through the above equation would involve a
square root operator. In order to preserve linearity, which is convenient for solving optimisation problems, the method presented in \cite{hybrid} is employed. It is possible to prove that a geometric mean inequality constraint written as
\begin{equation}
\sqrt{x_1 \cdot x_2} \geq x_3 \quad x_1 \cdot x_2 \geq 0,
\end{equation}
where $x_i \in \mathbb{R}$, can be reformulated as a second-order conic constraint
\begin{equation}
\begin{Vmatrix} 2 \cdot x_3 \\ x_1 - x_2 \end{Vmatrix}_2 \leq x_1 + x_2.
\end{equation}
Relaxing \eqref{Kinetic_Energy} as
\begin{equation}
E_{kin}(k) \geq \frac{1}{2} m v^2(k),
\end{equation}
and taking the square root from both sides we obtain
\begin{equation}
\sqrt{\dfrac{2E_{kin}(k)}{m}} \geq v(k).
\end{equation}
Hence, it is possible to reformulate the relaxed constraint as the following convex quadratic constraint
\begin{equation}
\begin{Vmatrix}
2 \cdot v(k+1) \\[1ex] \dfrac{2E_{kin}(k+1)}{m} - 1
\end{Vmatrix}_2 \leq \dfrac{2E_{kin}(k+1)}{m} + 1,
\label{eq: Ekin relaxed}
\end{equation}
linking the speed and the kinetic energy. The inverse of speed is usually defined as ‘lethargy’,
\begin{equation}
\frac{dt}{ds}(k) = \frac{1}{v(k)},
\label{eq:lethargy}
\end{equation}
that is the spatial derivative of time. Since \eqref{eq:lethargy} is a nonlinear constraint, we relax it and transform into a convex quadratic constraint
\begin{equation}
\begin{Vmatrix}
2 \\ \frac{dt}{ds}(k) - v(k)
\end{Vmatrix}_2 \leq \frac{dt}{ds}(k) + v(k).
\label{eq: lethargy relaxed}
\end{equation}
Finally, the lap time $t_{lap}$ can be expressed as
\begin{equation}
t_{lap} = \Delta s \sum_{i=0}^{N} \frac{dt}{ds}(k).
\end{equation}
Therefore, the relationship between the control inputs $\dot{v}(k)$, $k=0,...,N-1$ and the objective $t_{lap}$ is fully described by \eqref{eq: Ekin difference}, \eqref{eq: Ekin relaxed} and \eqref{eq: lethargy relaxed}, resorting to the intermediate variables $E_{kin}(k),\,v(k)$ and $\frac{dt}{ds}(k)$. 

\subsubsection{Constraints} To make the constraints of \eqref{eq:minimization} explicit, it is necessary to calculate the consumed fuel and used/recovered electrical energy in each portion of the track. The technical regulation defines the maximum fuel consumption per second at full thrust, that is $p^{max/s} = 0.0223 \, \frac{\text{kg}}{\text{s}}$. Given the combustion engine torque $T_{comb}(k)$, 
the amount of consumed fuel kilograms can be computed as
\begin{equation}
p(k) = p^{max/s} \cdot \frac{T_{comb}(k)}{T_{comb}^{max}(k)} \cdot \frac{dt}{ds}(k) \cdot \Delta s.
\end{equation}
From the thrust provided by the electric motor and its efficiency in traction $\eta_{el, traction}$, it is straightforward to compute the used electrical energy usage as
\begin{equation}
E_{el,used}(k) = \frac{T_{el}(k)}{\eta_{el, traction}} \cdot \Delta s. 
\end{equation}

The electrical energy can be recovered through the KERS both during sailing or in the braking phase. In the first case, the KERS intervenes to partly recover the kinetic energy ($F_{sail}\leq 0$). In the second case, instead, the energy is recovered through a braking force $F_{dec} \leq 0$, lower in module than the maximum value $F_{dec,max} \leq 0$ that can be guaranteed by the brakes and the adherence with ground. Hence, according to the situation, the recovered electrical energy can be estimated using one of the two expressions
\begin{subequations}
\begin{equation}
E_{el, rec-KERS}(k) = |F_{sail}|(k) \cdot \Delta s \cdot \eta_{el, rec},
\end{equation}
\begin{equation}
E_{el,rec-KERS}(k) = \min(|F_{dec}|(k), |F_{dec,max}|) \cdot \Delta s \cdot \eta_{el, rec}, 
\end{equation} 
\end{subequations}
where $\eta_{el, rec}$ is the efficiency of the electric motor in recuperation phase. The constraints of \eqref{eq:minimization} can be finally reformulated as
\begin{subequations}
\begin{equation}
\sum_{s=0}^{N} p(k) \leq p^{max} 
\end{equation}
\begin{equation}
\sum_{s=0}^{N} E_{el,used}(k) \leq E_{el,used}^{max}
\end{equation}
\begin{equation}
\sum_{s=0}^{N} E_{el,rec-KERS}(k) \geq \sum_{s=0}^{N} E_{el,used}(k) - E_{el,rec-HERS} 
\end{equation}
\end{subequations}

\begin{figure}[htb]
	\centerline{\includegraphics[width=\columnwidth]{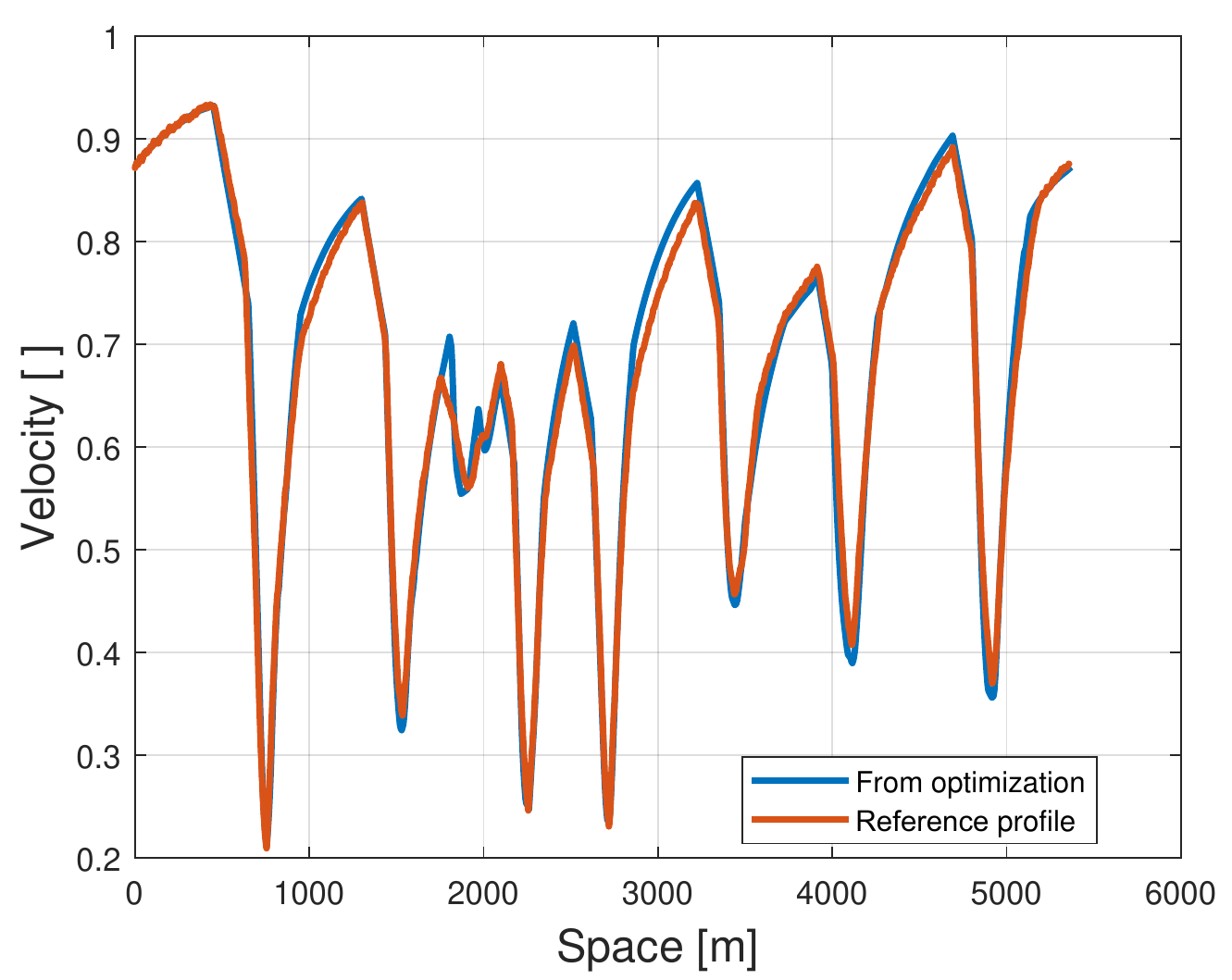}}
	\caption{Comparison between the reference speed profile and the optimal one, obtained by solving the MIQCP.}
	\label{fig:speed_comp}
\end{figure}

\begin{figure}[htb]
	\centerline{\includegraphics[width=\columnwidth]{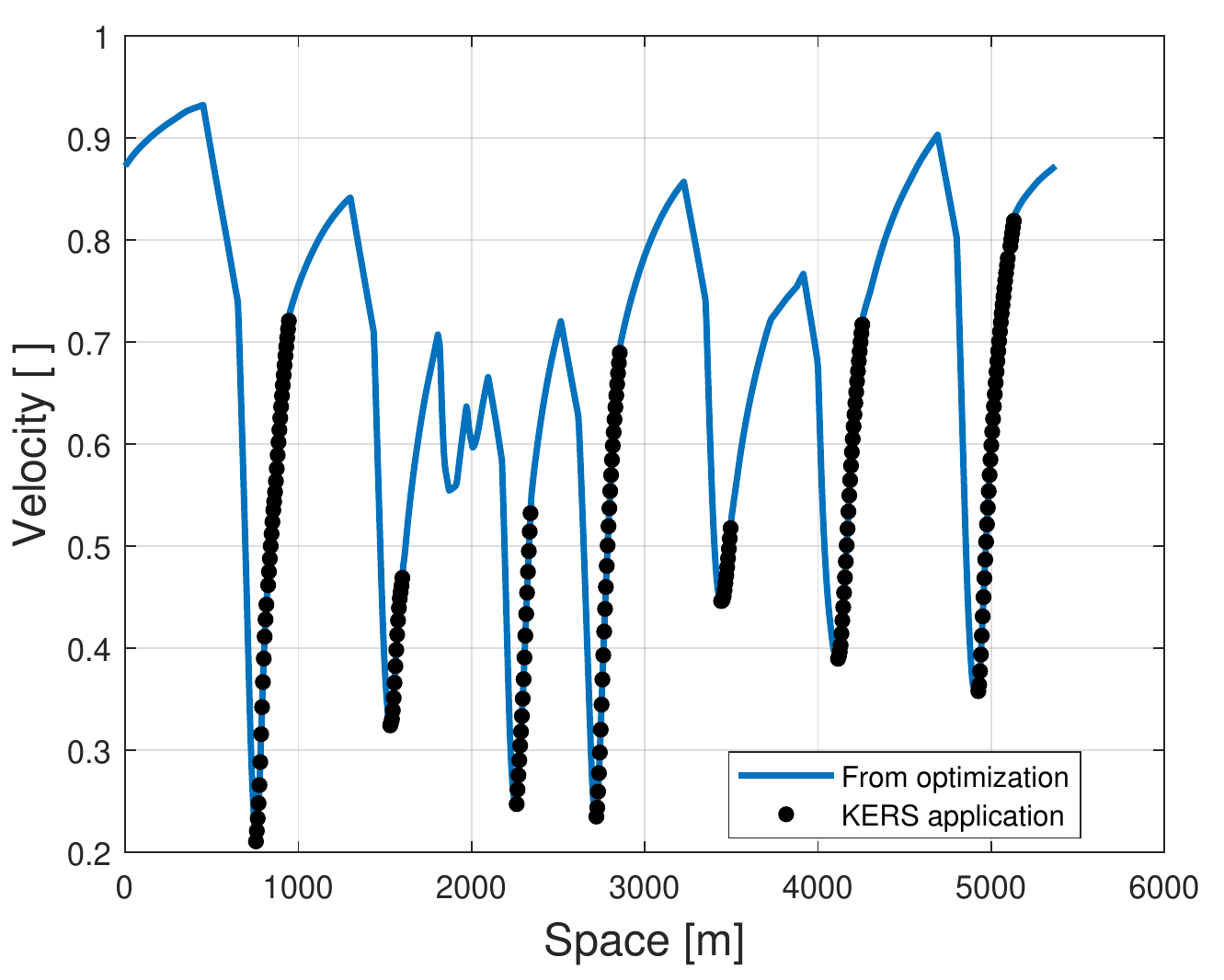}}
	\caption{MIQCP-based optimal speed profile with detail of the electric motor activation points.}
	\label{fig:speed_KERS}
\end{figure}

The optimisation problem was built using the toolbox YALMIP \cite{yalmip1}. The resulting Mixed Integer Quadratically Constrained Program (MIQCP) was then solved with IBM\textsuperscript{\textregistered} ILOG\textsuperscript{\textregistered} CPLEX\textsuperscript{\textregistered} version 12.8.0.  The optimised speed profile is shown in Figure \ref{fig:speed_comp}, as compared with the reference one. There is significant coherence between the reference speed profile and the one resulting from optimisation. Finally, the points of activation of the electric motor are highlighted in Figure \ref{fig:speed_KERS}, from which we can conclude that the solver decides to use the electric motor in the first part of the straights, achieving the highest possible speed in the shortest time. This is coherent with the behaviour of professional human drivers.

\subsection{Genetic-Algorithms-based solver}
\label{sec: Genetic Algorithm}
As previously mentioned, we now propose an alternative resolution method for the optimisation problem based on Genetic Algorithms. 
The circuit is divided into $M=8$ regions, each one from the beginning of a straight to the beginning of the following one. Differently from the MIQCP approach, whose optimisation variable embodies the powertrain usage mode directly, we consider $2$ optimisation variables for each region, that is the electrical energy consumption $E_{el, used}$ and the fuel consumption $p$. The total number of optimisation variables in a lap is therefore $2M=16$. We further assume that the energy budget of each region, encoded by the optimisation variables, is actuated continuously from the beginning of the region (straight) until it is completely consumed. This assumption is supported by the fact that the optimal way to apply energy is to use it in the first part of the acceleration region, as known to the experts of the field. Finally, we highlight that, given an instance of the optimisation variables and the previous assumption, it is possible to uniquely identify the corresponding powertrain usage mode, speed profile and lap time.  

With reference to the traditional nomenclature of GA, an instance of the $2M$ optimisation variables constitutes the genome of an individual of the population. We employ a discretisation step of $1$ kJ for the electrical energy consumption and $1$ g for the fuel consumption, and select the optimisation variables ranges as $\left[ 0, 2300 \right]$ $\frac{\text{kJ}}{\text{region}}$ and $\left[ 0, 1381 \right]$ $\frac{\text{g}}{\text{region}}$, respectively. The discretisation speeds up mutations and crossovers to reach the optimised solution, and constitutes a fair approximation being the step sizes three orders of magnitude lower than the maximum value that the variables can take. The inequality constraints \eqref{eq: minimization 3} and \eqref{eq: minimization 4} are then enforced through linear inequalities on the optimisation variables.
Finally, the fitness function is the lap time generated by the genome.

A reasonable initial population was provided to the GA to hot start the optimisation and improve computational times. The simplest way to obtain a hot-start genome in the population is to compute the amount of electrical energy and fuel consumed in each region of the circuit in a real lap. The chosen reference lap was performed in absence of interactions with competitors and with the lowest lap time possible, so as to be close to the optimal solution. Another important hyperparameter to be carefully selected is the population size, which is critical for the convergence of the GA. Literature provides some heuristic methods to select the dimension of the population. Anyway, for the majority of the problems they indicate that an increase in the population size statistically reduces the error between the solution found by the GA and the exact solution \cite{GA1},
at the expense of an increase in the computational time. The most common rule of thumb establishes that the dimension of the population needs to be at least equal to the size of the genome. Converting the variables from decimal to binary, as typical for GA, we obtain
\begin{equation}
\begin{split}
(2300)_{10} = (100011111100)_2 \, &\rightarrow \, 12 \, \text{bit},\\
(1381)_{10} = (10101100101)_2 \, &\rightarrow \, 11 \, \text{bit}, 
\end{split}
\end{equation}
from which the initial population needs to have at least $23M = 184$ individuals. After analysis of the results, we finally changed the population size to $2000$ individuals to obtain stable and repeatable solutions. 

\begin{figure}[tbp]
\centering
	\subfloat[Strategy 1.]{\includegraphics[width=0.48\columnwidth]{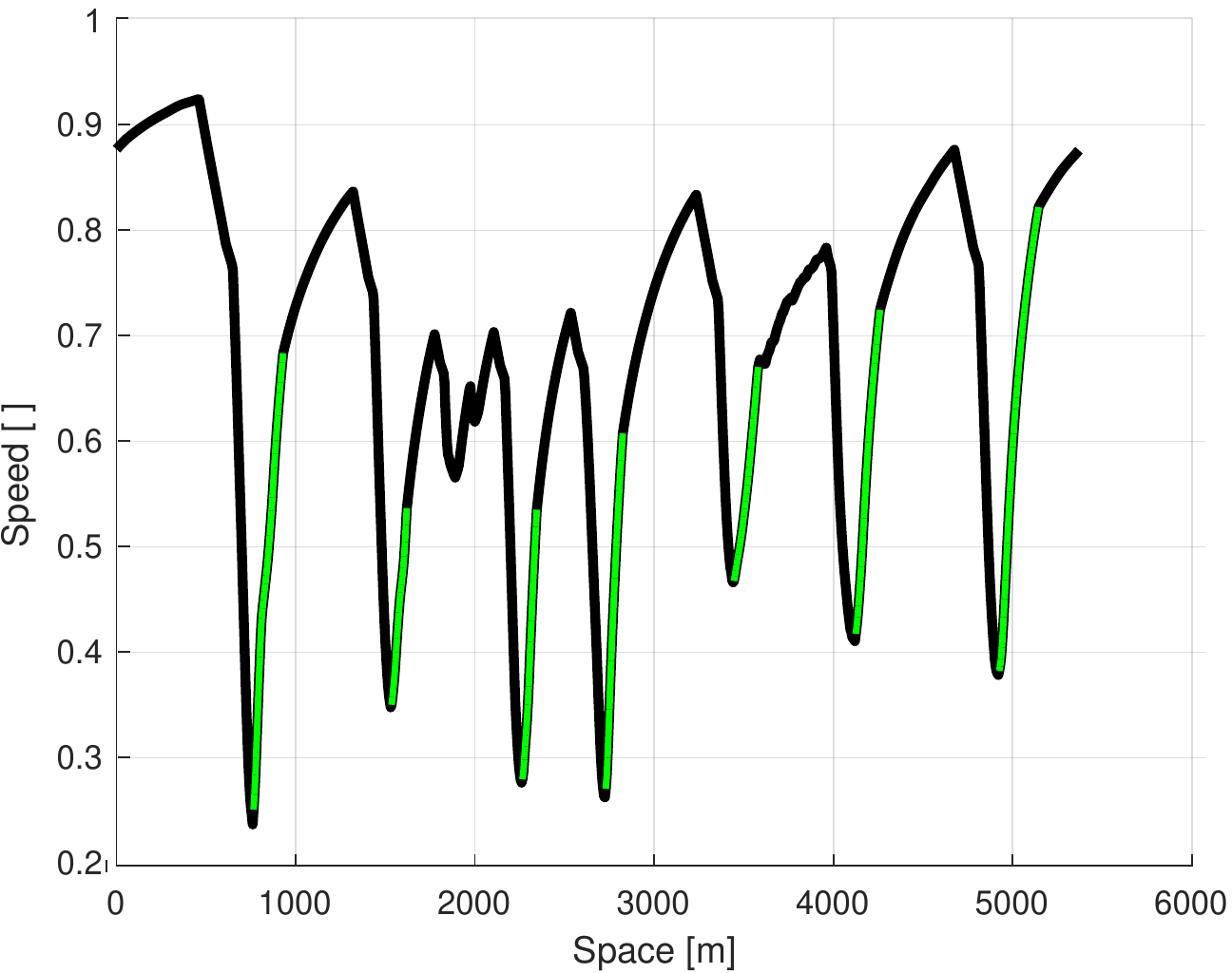}}
	\subfloat[Strategy 5.]{\includegraphics[width=0.48\columnwidth]{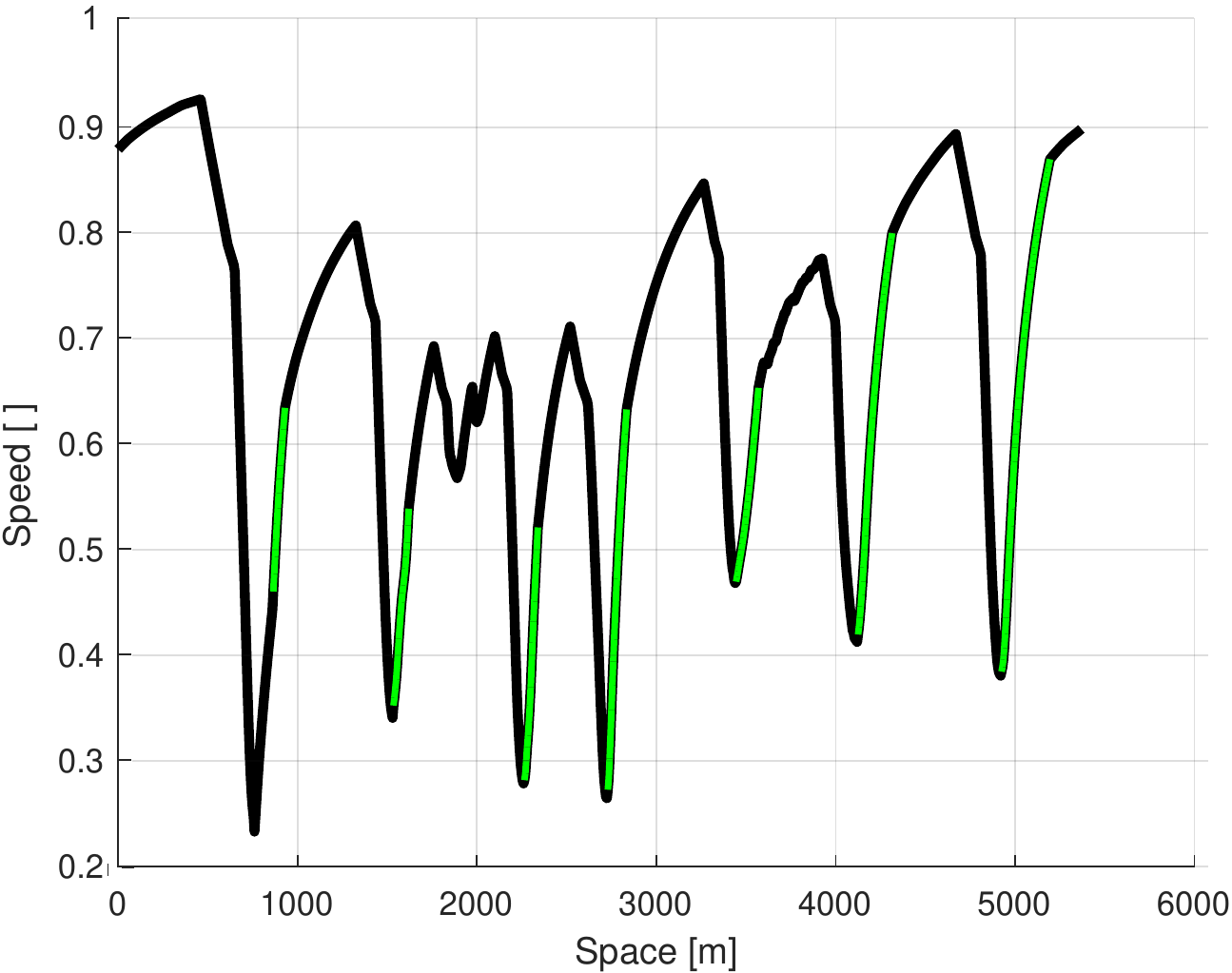}}\\
	\subfloat[Strategy 13.]{\includegraphics[width=0.48\columnwidth]{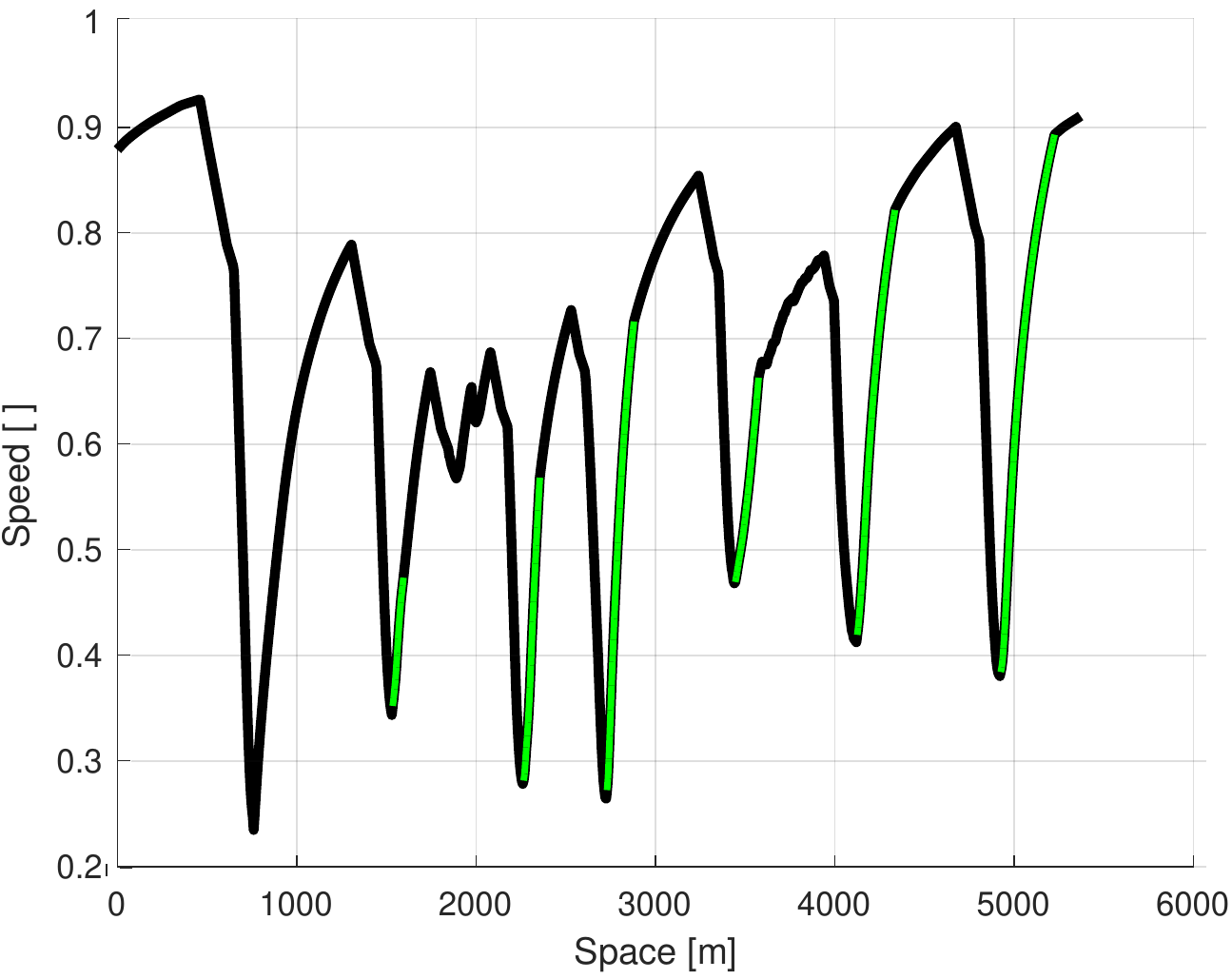}}
	\subfloat[Strategy 10.]{\includegraphics[width=0.48\columnwidth]{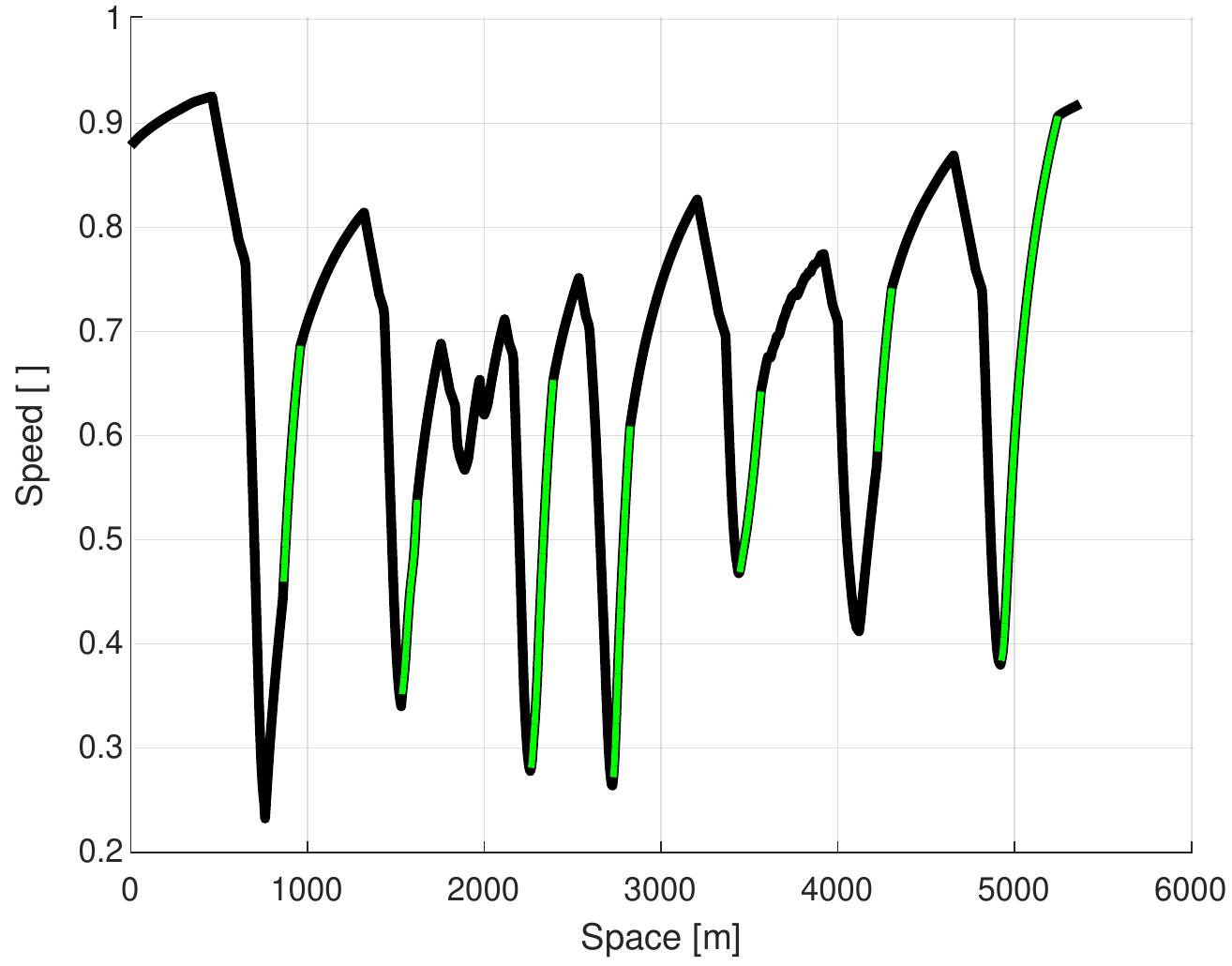}}
	\caption{Speed profiles resulting from the GA-based optimisation. The green lines represent the points of activation of the electric motor.}
	\label{fig:speed_GA}
\end{figure}

Differently from the previous method based on the MIQCP formulation, which generates a single optimal solution in absence of competitors, we employ the GA to create a set of optimal energy strategies, whose efficacy will be then evaluated in realistic traffic conditions. The evaluation is performed in the next section. The set of strategies is obtained by launching the GA many times, each with different constraints on the electrical energy usage. The underlying reasoning for heuristically imposing extra constraints is the following. Since the strategies will be evaluated in complex traffic scenarios, activating the electric motor where overtaking is easier to be performed, e.g. a straight, can be beneficial to reduce the time loss related to the maneuver and the lap time. Table \ref{tab:policiesGA} shows the constraints applied in each strategy and the corresponding optimal lap time. It can be noticed that the applied constraints have an action range of at least 100 meters. This is done to respect the sensitivity/capability of the driver to manually intervene on the power-off/on button of the electric motor.
Some exemplary optimal strategies are shown in Figure \ref{fig:speed_GA}.

\begin{table*}[htb] \normalsize
	\caption{Strategies and their Electrical Energy constraints generated by the Genetic Algorithm}
	\label{tab:policiesGA}
	\centering
	
	\begin{tabular}{ccc}
		\toprule
		{\bfseries Strategies \#} & {\bfseries Lap time} (s) & {\bfseries Electrical Energy constraints}\\
		
		\midrule
		1	& 104.028 & No constraints \\
		2	& 104.092 & No KERS first 100 meters $8^{th}$ straight \\
		3	& 104.152 & No KERS first 100 meters $2^{nd}$ straight \\
		4	& 104.273 & No KERS first 110 meters $4^{th}$ straight \\
		5	& 104.290 & No KERS first 100 meters $1^{st}$ straight \\
		6	& 104.310 & No KERS first 200 meters $8^{th}$ straight \\
		7	& 104.373 & No KERS first 100 meters $7^{th}$ straight \\
		8	& 104.396 & No KERS first 300 meters $8^{th}$ straight \\
		9	& 104.402 & No KERS first 100 meters $7^{th}$ and $8^{th}$ straights \\
		10	& 104.415 & No KERS first 100 meters $1^{st}$ and $7^{th}$ straights \\
		11	& 104.417 & No KERS first 200 meters $7^{th}$ straight \\
		12	& 104.566 & No KERS first 100 meters $1^{st}$ and $8^{th}$ straights \\
		13	& 104.603 & No KERS first 200 meters $1^{st}$ straight \\
		14	& 104.674 & No KERS first 70 meters $5^{th}$ straight \\
		15	& 104.733 & No KERS first 140 meters $5^{th}$ straight \\
		\bottomrule
	\end{tabular}	
\end{table*}

\section{Multi-agent simulations of the competitors' behaviour}
\label{sec:on_simulations}
The approach presented in Sec. \ref{sec:off_lap_optimization} generated a set of offline strategies for the lap time minimisation in absence of competitors. To evaluate them in realistic race conditions, where competitors are present along the track and mutually interact with each other, it is necessary to simulate their behaviour. In this section, we explain how to build multi-agent Monte Carlo simulations of the competitors' behaviour from the statistics computed in Sec. \ref{sec:competitors_performance}. 

\begin{figure}[tbp]
	\centerline{\includegraphics[width=0.95\columnwidth]{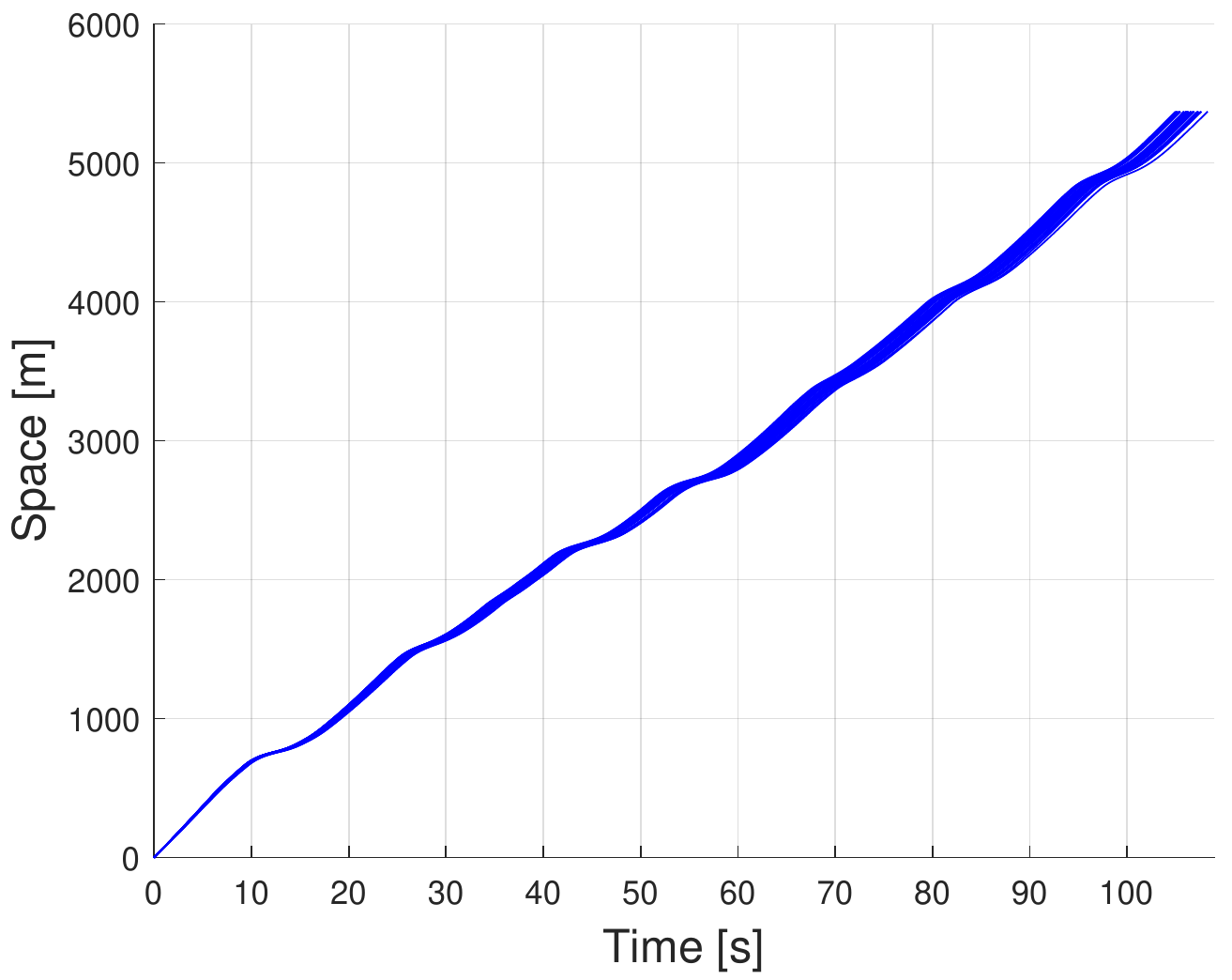}}
	\caption{Simulated distribution of forecast positions of an exemplary competitor in absence of mutual interactions.}
	\label{fig:free_sector2}
\end{figure}

The forecasts have to consider a time horizon that is long enough to allow the ego-car to complete the lap. For this reason, we decide to perform two-laps-long forecasts starting from the initial condition. 
The positions of the competitors in absence of mutual interactions are forecast by means of the free sector times probabilistic distributions, derived from Sec. \ref{sec:competitors_performance}.
With reference to Figure \ref{fig:FreeSectorTimes}, we remark that the associated probability distributions are not Gaussian.
To cope with this, we perform random sampling, with enough extractions to guarantee that the probabilistic distributions of the competitors' behaviour can be reliably approximated.
The forecast position along the track of an exemplary competitor for different simulations is shown in Figure \ref{fig:free_sector2}.

At this point, we introduce the interactions between the competitors.
To do so, a multi-agent-based model has been developed, resorting to the Influence/Reaction principle \cite{InfluenceReaction, Agent1, Agent2}. Each competitor is considered as a rational agent that tries to minimise its lap time. If a competitor gets closer to the preceding one, it perceives its influence and has to react, performing overtaking or following.
When an overtaking may be performed, the category of the involved vehicles is evaluated and, taking into account the section along the circuit where the overtaking may occur, the overtaking probability is computed from the statistics derived in Sec. \ref{sec:competitors_performance}.
A random number in the range $[0,1]$ is extracted and compared to the overtaking probability. If it is lower, then overtaking succeeds, otherwise the car keeps on following the preceding car for the entire length of the section. The procedure is then repeated at the successive section. 

\begin{figure}[htb]
	\subfloat[Example 1.]{\includegraphics[width=0.95\columnwidth]{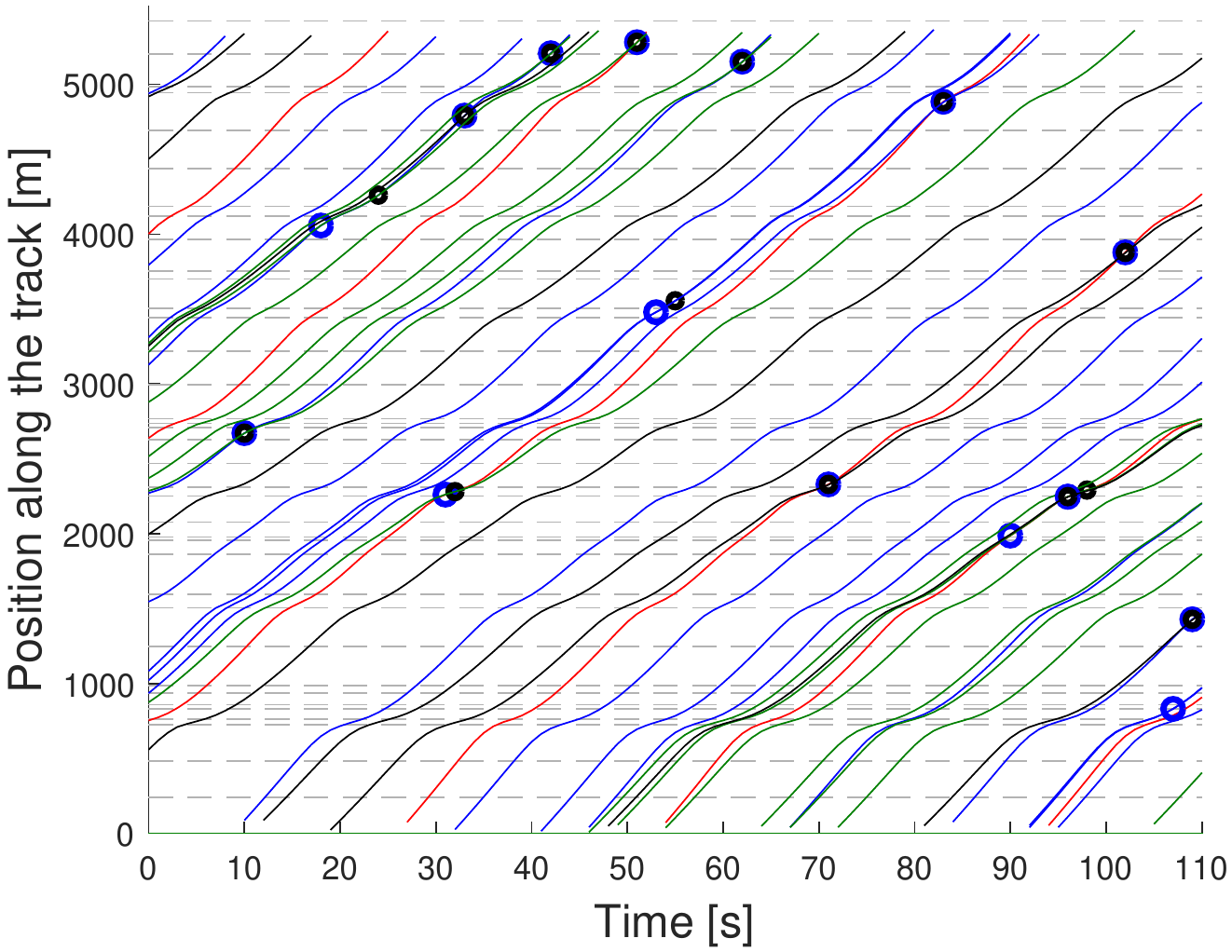}}\\
	\subfloat[Example 2.]{\includegraphics[width=0.95\columnwidth]{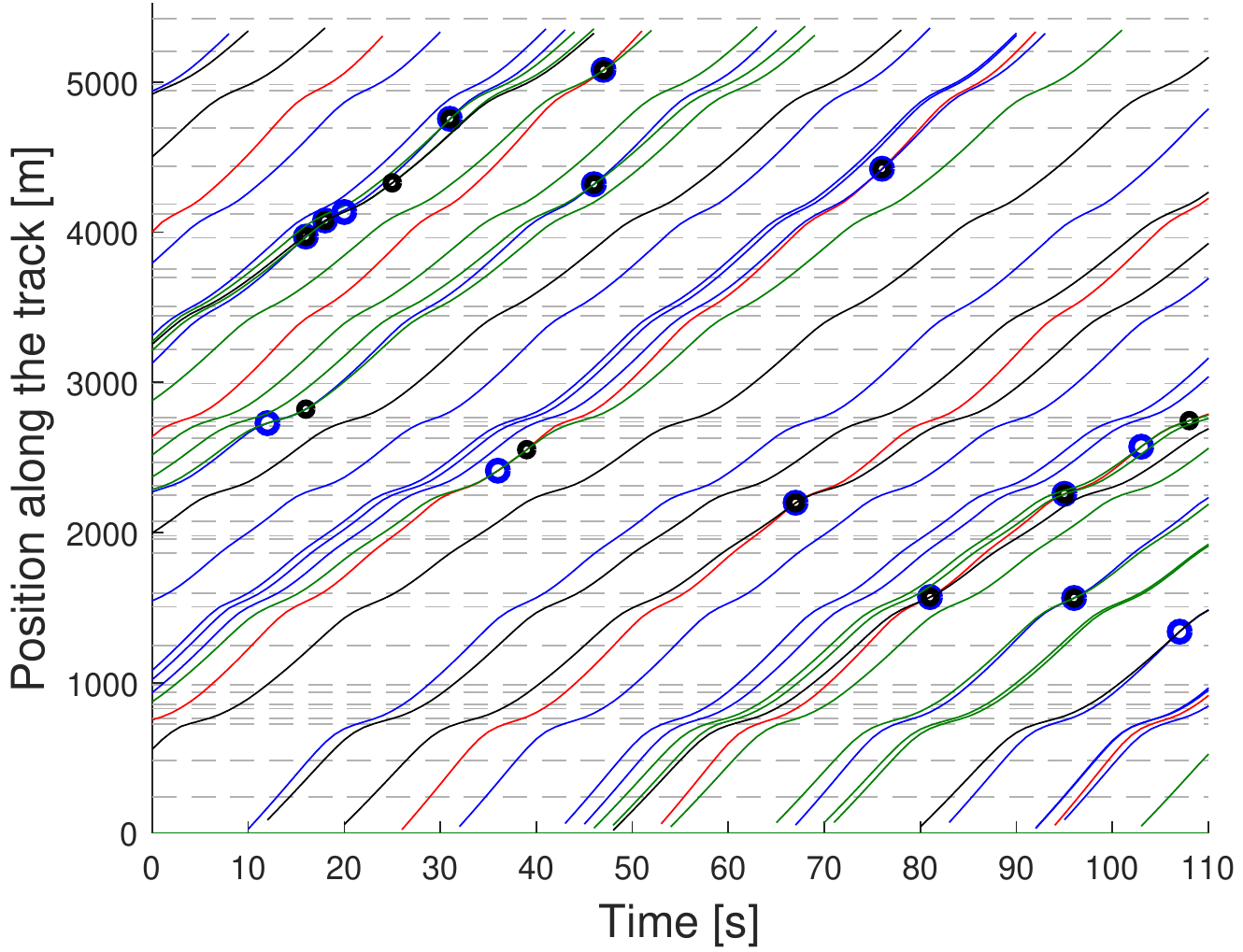}}
	\caption{Monte Carlo numerical simulations of the competitors' motion along the track. The colours of the lines represent different categories: red for LMP1, blue for LMP2, green for LMGTE Pro and finally black for LMGTE Am. The circles, instead, have the following meaning: the blue one represents the position along the circuit where two vehicles get close enough to activate the Influence/Reaction principle, whereas the black one defines the position where the overtaking occurs.}
	\label{fig:MC}
\end{figure}

Repeating these steps for all possible overtakings involving the competitors along the circuit generates a Monte Carlo numerical simulation. Two examples of Monte Carlo simulation are reported in Figure \ref{fig:MC}. 

To reliably forecast the behaviour of the competitors, it is necessary to generate a significant number of Monte Carlo numerical simulations. This means performing a large amount of random samplings of free sector times and overtaking probabilities.
These computations may however be time consuming. Indeed, they have to be performed in real time at the beginning of each lap, as they depend on the actual positions of the competitors along the circuit. A solution to this issue can be addressed through parallel computing, since the simulations are completely independent from each other. The statistical analysis of the outputs and evaluation of the optimal strategy in presence of realistic traffic conditions is described in the next section.

\section{Stochastic lap strategy optimisation in traffic conditions}
\label{sec:results}
Optimal strategies for the powertrain energy budget in absence of traffic have been presented in Sec. \ref{sec:off_lap_optimization}. On the contrary, the Monte Carlo approach defined in \ref{sec:on_simulations} allows to forecast the positions of the competitors, accounting for their mutual interactions. We now propose a novel way to combine the two features and build a stochastic optimal solver, aiming to identify the strategy 
that statistically guarantees the minimum lap time in presence of traffic.  

Stochastic Dynamic Programming (SDP), which is the method we employ in this work, is an optimisation-based method for making sequential decisions under uncertainty. 
In our framework, uncertainty is associated to the occurrence of overtakings. Being SDP a discrete decision-making process and considering $n$ stages $t = 0, \dots, n-1$, the following ingredients are defined at each timestep $t$:
\begin{itemize}
	\item an initial state $s_t \in \mathbb{S}_t$, where $\mathbb{S}_t$ is the set of feasible states at stage $t$;
	\item a decision variable $x_t \in \mathbb{X}_t$, where $\mathbb{X}_t$ is the set of feasible actions that can be chosen at stage $t$;
	\item an immediate cost or reward function $r_t (s_t, x_t)$, representing the incurred cost or the gained reward at stage $t$ if an action $x_t$ is chosen at the state $s_t$;
	\item a state transition function $g_t(s_t, x_t)$, defining the state change from $s_t$ towards $s_{t+1}$;
	\item a discount factor $\alpha \in [0,1]$;
	\item a conditional probability $Pr(s_{t+1} \lvert s_t,x_t)$, representing the probability to move into state $s_{t+1}$ given the current state $s_t$ and the chosen action $x_t$.
\end{itemize}
The optimal control problem can be iteratively described through the value function $f_t$ at the generic stage $t$ as
\begin{equation} \footnotesize
f_t(s_t) = \min_{x_t \in \mathbb{X}_t(s_t)} \biggl(r_t(s_t,x_t) + \alpha \sum_{s_{t+1}} Pr(s_{t+1} \lvert s_t,x_t) f_{t+1}(s_{t+1}) \biggr),
\label{eq:SDP}
\end{equation}    
which represents the expected optimal cost that can be attained from the state $s_t$ if the optimal action $x_t$ is chosen. 

According to \eqref{eq:SDP}, the optimal strategy is the one that minimises the value function at the initial stage of the problem, and can be computed resorting to policy iteration. However, we have already selected fifteen strategies a priori in Sec. \ref{sec:off_lap_optimization}, through the energy budget optimisation problem. Thus, we do not perform policy iteration, but policy evaluation: the value function deriving from each of the pre-computed strategies is evaluated. Finally, the best strategy is identified as the one that minimises the value function at the first stage. 

\begin{figure}[tbp]
	\centerline{\includegraphics[width=0.95\columnwidth]{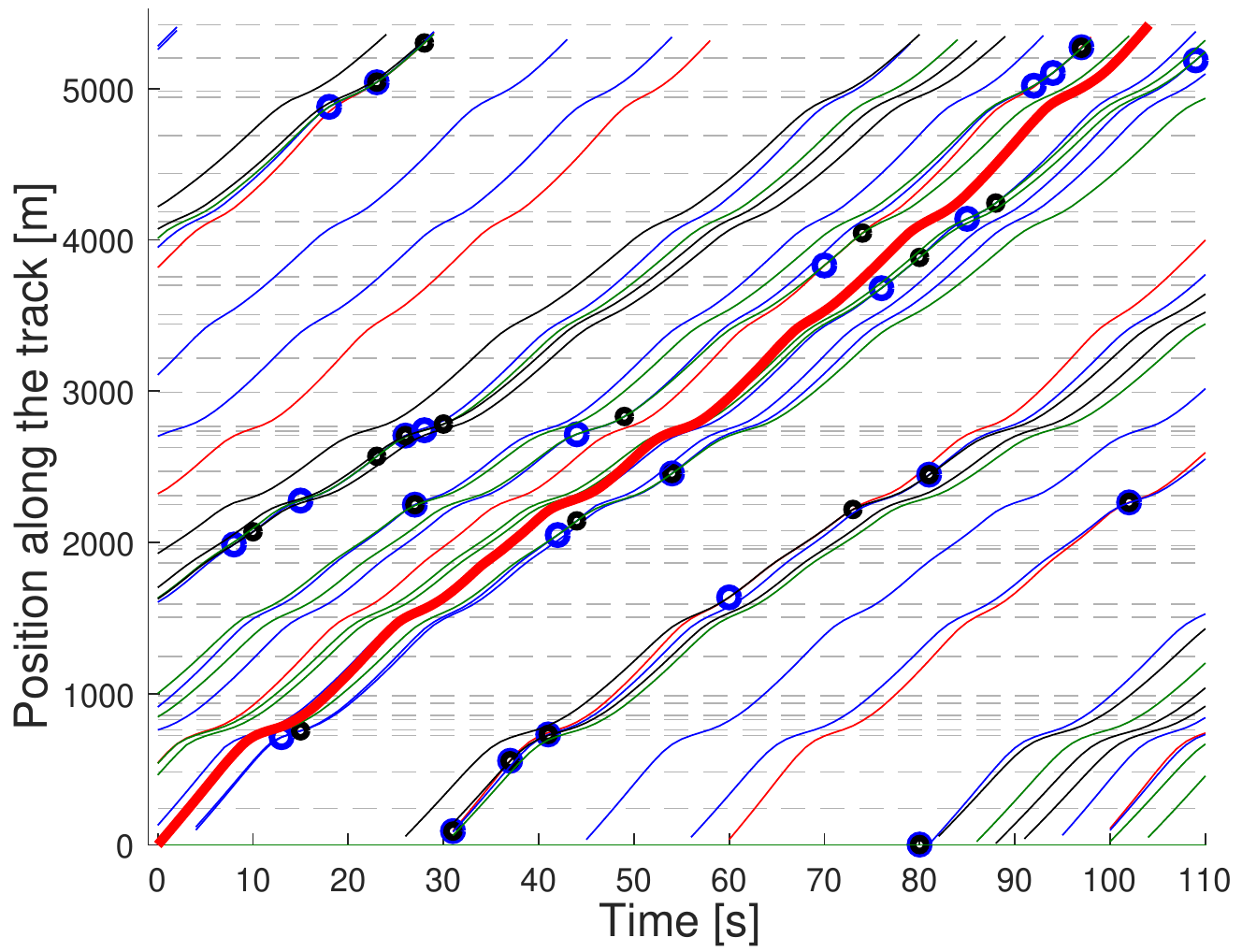}}
	\caption{Testing the energy budget strategy in a Monte Carlo numerical simulation. The thick red line represents the spatial profile of the reference car, whereas the other lines represent the competitors.}
	\label{fig:MC+Policy}
\end{figure}

To apply this technique, we first test each of the fifteen optimal energy strategies in each of the Monte Carlo numerical simulations. Figure \ref{fig:MC+Policy} reports an example of this procedure. 
The parameters of this technique have been defined in the following way. The variable $s$, representing the state of the system, consists in the negative number of the possible overtakings to be performed in the reference lap. Indeed, the positions of the reference car and of the competitors are known from respectively the optimisation problem and the Monte Carlo numerical simulations. Thus, for each combination of optimal strategies and numerical simulations it is possible to compute the number of possible overtakings that the reference car can perform in the lap. For instance, with reference to Figure \ref{fig:MC+Policy}, three overtakings can occur. This implies that the state of the system is initialised to $s_1 = -3$. When an overtaking occurs, the state is augmented by 1. The decision variable $x_t$ consists in the usage of the powertrain energy budget, either coming from the combustion engine and from the electric motor. The variable $Pr$ refers to the probability of changing the state from $s_t$ to $s_{t+1}$, once a specific decision variable $x_t$ is chosen. Since we have defined the state as the number of possible overtakings to be performed in a lap, this probability is exactly the overtaking probability that we have drawn from the analysis of previous years events as a function of the category of the involved vehicles and of the section where the overtaking may occur.

The variable $r_t$ is defined as a cost. When the reference car approaches a competitor, according to the categories and the section where the vehicles are, there is a specific overtaking probability. 
If overtaking is not performed, the ego-vehicle needs to slow down, which involves a time loss. The latter can be computed as the difference between the time instant at which the reference car has reached the end of the section behind the preceding car and the instant of time at which the reference car would have reached the end of the section without traffic. Moreover, if the reference car is obliged to stay behind the preceding car in a section, it has to reduce its speed.
Once it is able to overtake the slower vehicle in a successive section, it needs time before reaching its reference optimal speed profile again. This time loss is also added to the cost.

The parameter $\alpha$ is a factor that discounts future costs, which is typically employed in infinite-time problems. 
Since we deal with a finite-time problem, the parameter is set equal to $1$. 

\begin{figure}[tbp]
	\subfloat[Probability Decision Tree. The number above the state represents the probability of occurrence of a particular event.]{\includegraphics[width=0.95\columnwidth]{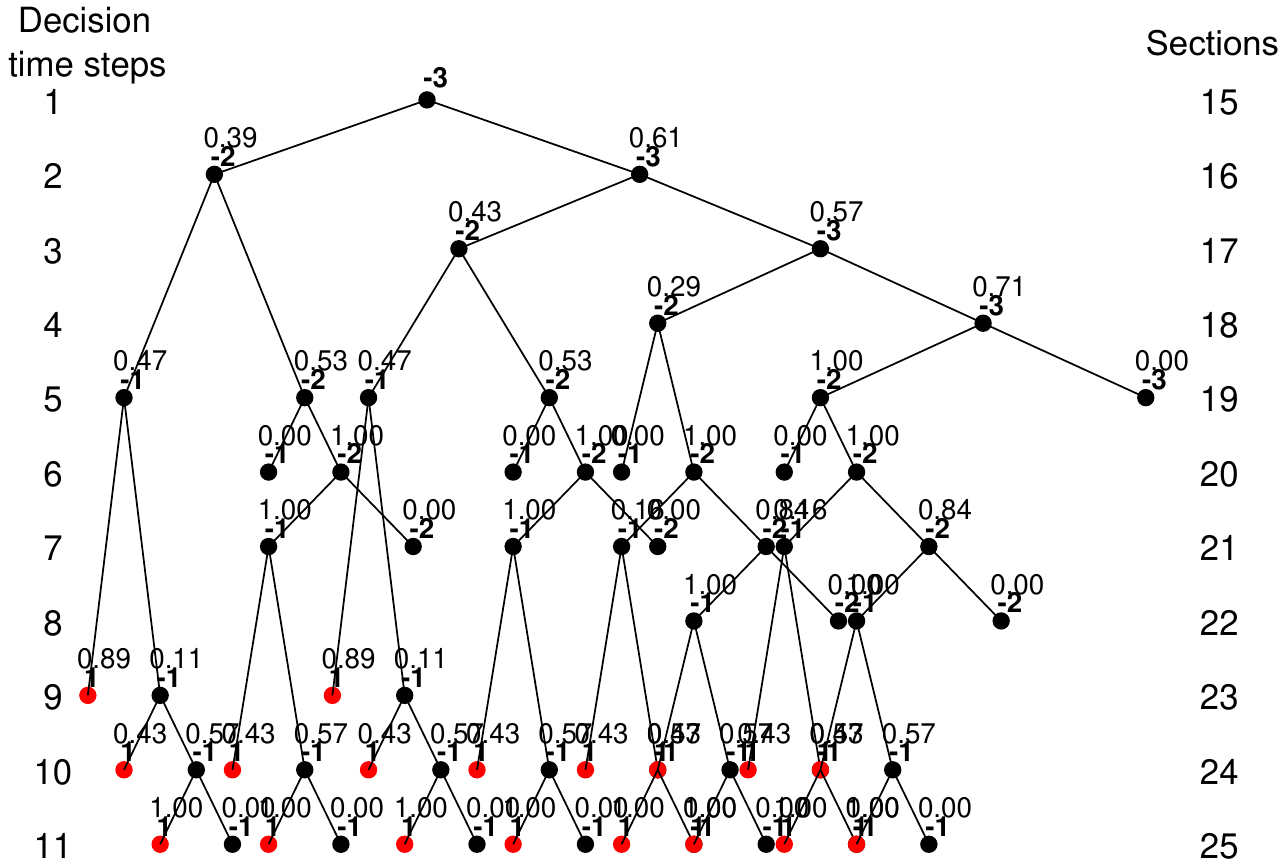}}\\
	\subfloat[Cost Decision Tree. The number above the state represents the cost of a particular event, i.e. its associated time loss expressed in seconds.]{\includegraphics[width=0.95\columnwidth]{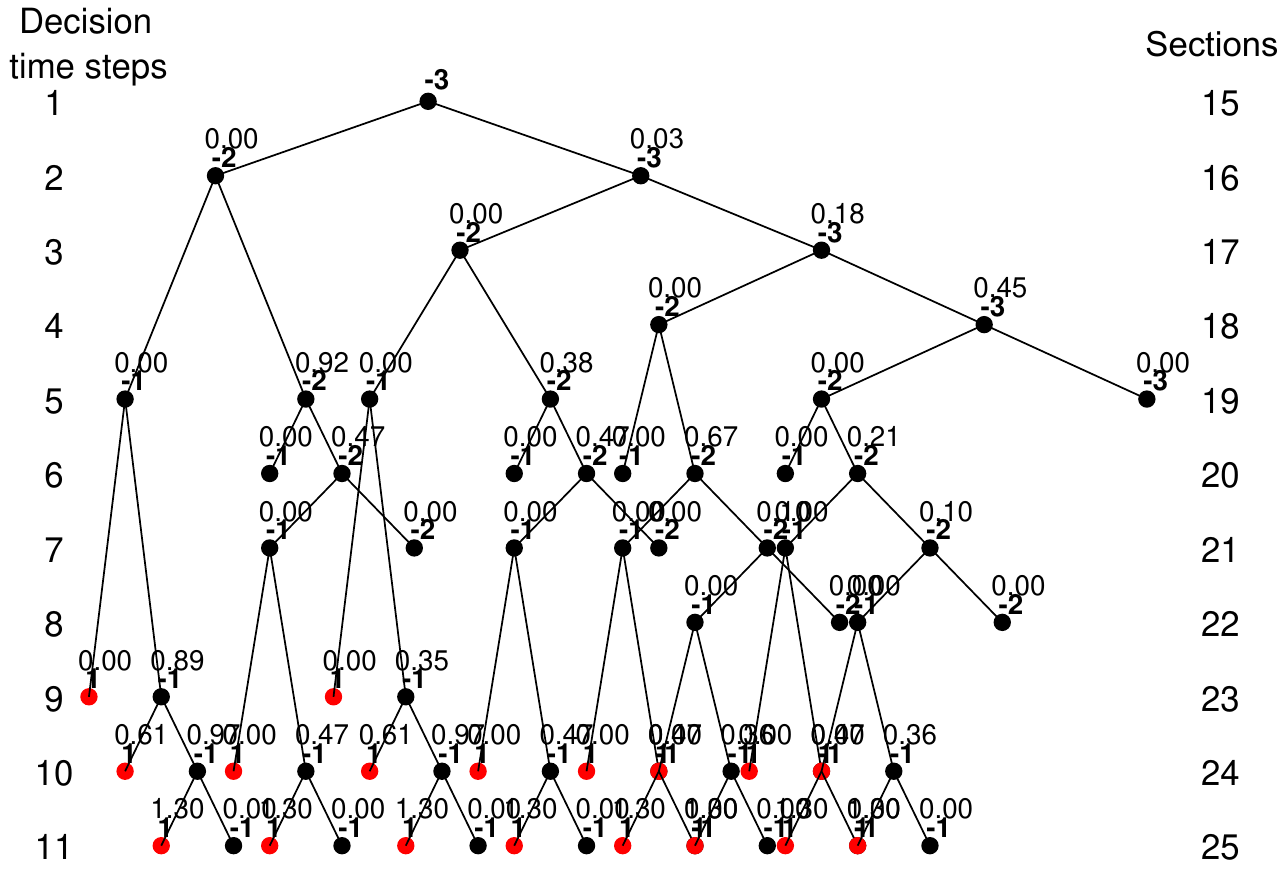}}
	\caption{Example of Probability and Cost Decision Trees. The negative value over each circle represents the state of the system, i.e. the number of overtakings still to be performed in the lap. A red circle with state equal to 1 indicates that all possible overtakings have already been performed for that probabilistic event. The left column of numbers identifies the decision stages where overtakings may occur, whereas the right column specifies the section along the track.}
	\label{fig:tree}
\end{figure}

The $n$ stages of SDP coincide with the $37$ sections defined for the computation of the overtaking probabilities in Sec. \ref{sec:competitors_performance}. In each section the algorithm evaluates whether the reference car may perform an overtaking. If a competitor is close enough, there is a specific probability of performing the overtaking. 
If overtaking succeeds, the state is augmented by one and there is no associated time loss. On the contrary, the reference car is obliged to follow the preceding vehicle, the state does not change, and a time loss is introduced in the cost function
It is convenient to graphically represent all of the probabilistic events rising from overtakings in a lap. In this regard, we adopt a suitable and common representation called Decision Tree. To build Decision Trees, the possible occurrence of an overtaking is evaluated in each section from the beginning to the end of the track (forward pass \cite{forward}). Then, a backward pass computes the value function at the initial stage of the problem, i.e. $f_0$. It consists in multiplying the probability of an event with the associated cost values, and summing them up stage by stage from the end to the beginning, according to \eqref{eq:SDP}. A portion of the Decision Trees for the optimal strategy is reported in Figure \ref{fig:tree}. 

\begin{figure}[htb]
 	\centerline{\includegraphics[width=0.7\columnwidth]{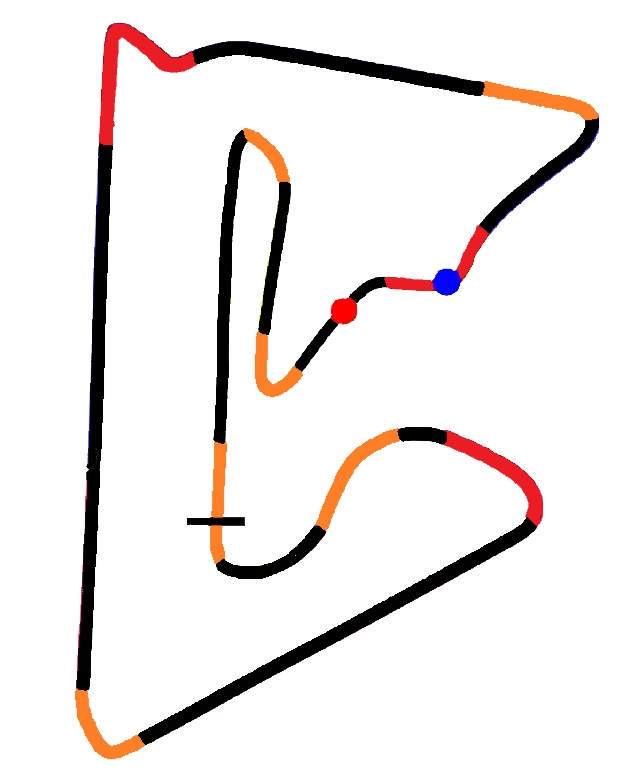}}
 	\caption{Analysis of time losses due to overtakings in a full lap. The black-orange-red colours of the sections represent increasing time losses due to overtakings. The blue circle marks the position where the first overtaking is probable to occur, according to the forecast of the competitors positions and to the default energy budget strategy, i.e. the first strategy. The red circle expresses the same concept for the fifth energy strategy, i.e. the optimal one. Since this strategy forces the solver not to use the electric motor in the first straight, the ego-car is expected to perform the first overtaking further along the track. This highlights that the proposed approach is suitable to shift overtakings to positions where they cause lower time losses.}
 	\label{fig:time_loss_track}
\end{figure}

\begin{figure}[htb]
	\centerline{\includegraphics[width=0.95\columnwidth]{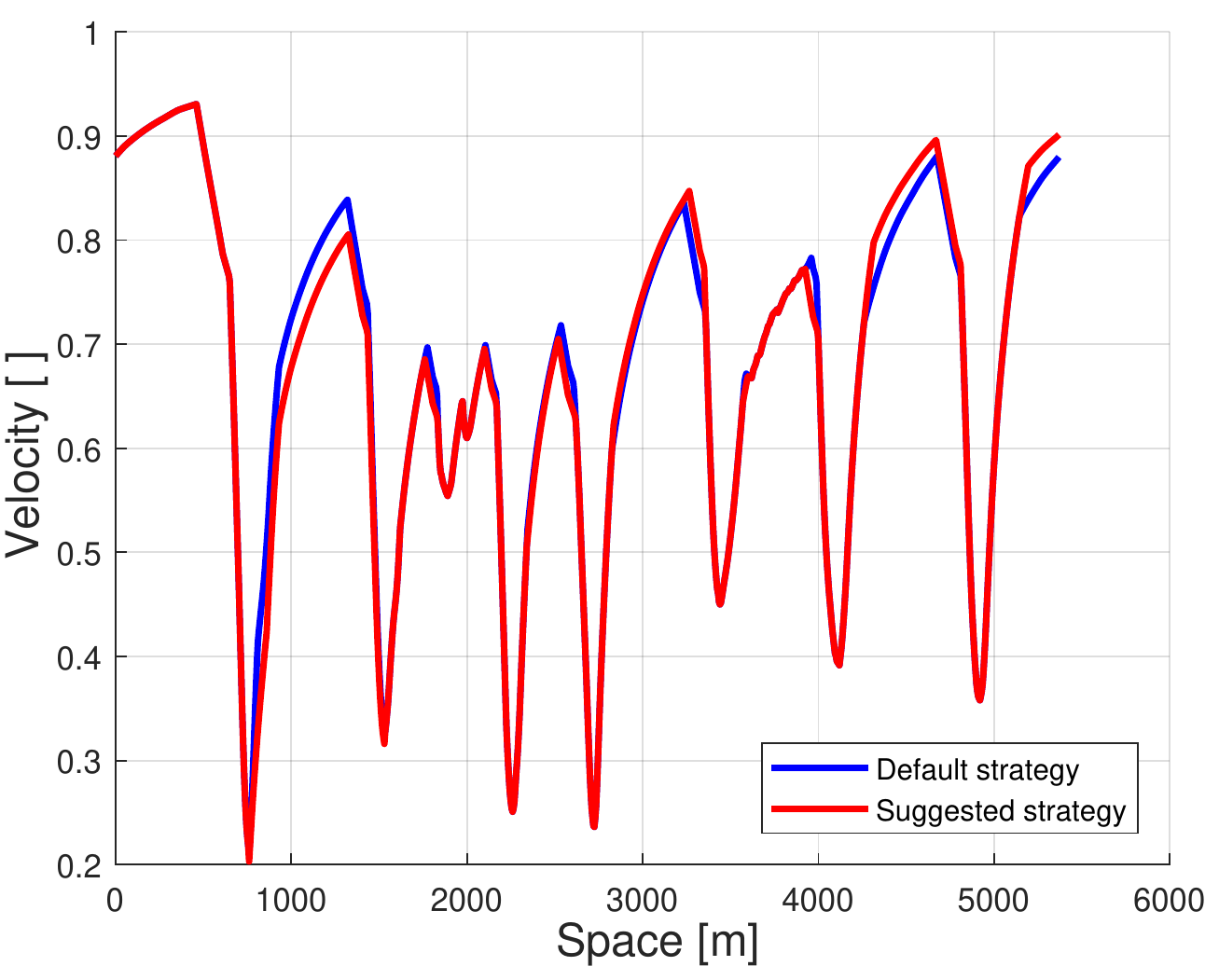}}
	\caption{Speed profiles resulting from the first and fifth strategy. The optimal strategy reaches a lower top speed in the first straight, due to a lower usage of the electrical energy. The saved energy is then spread into the last two straights, where it is easier to perform overtakings and reach higher speed.}
	\label{fig:strategies_speed_prof}
\end{figure}

The value functions at time zero $f_0$ of each strategy are reported in Table \ref{tab:f1_results}. 
The fifth strategy results to be the optimal one, having the lowest time loss. Despite this strategy is approximately 0.3 s slower than the first one in absence of traffic conditions, i.e. the reference car, it is statistically faster of approximately 0.5 s considering traffic conditions. Moreover, the fifth strategy has a high statistical significance, since it resulted to be the optimal strategy in approximately 70\% of the Monte Carlo numerical simulations. Further graphical interpretations are provided in Figures \ref{fig:time_loss_track} and \ref{fig:strategies_speed_prof} to support our approach. They both show that the optimal strategy tends to save electric power and release it to perform overtakings only in portions of the track where they produce lower time losses. 

\begin{table}[tbp]\normalsize
	\caption{Value function at the initial stage from Stochastic Dynamic Programming}
	\centering
	
	\begin{tabular}{cc}
		\toprule
		{\bfseries Strategy \#} & \bfseries $f_0$\\
		& (s)\\
		\midrule
		1 & 2.131 \\
		2 & 1.987 \\
		3 & 1.792 \\
		4 & 1.884 \\
		5 & 1.662 \\
		6 & 2.092 \\
		7 & 2.176 \\
		8 & 2.338 \\
		9 & 2.359 \\
		10 & 2.421 \\
		11 & 2.428 \\
		12 & 2.619 \\
		13 & 2.211 \\
		14 & 2.886 \\
		15 & 2.534 \\
		\bottomrule
	\end{tabular} 
	\label{tab:f1_results}
\end{table}

\begin{figure}[tbp]
	\centerline{\includegraphics[width=0.95\columnwidth]{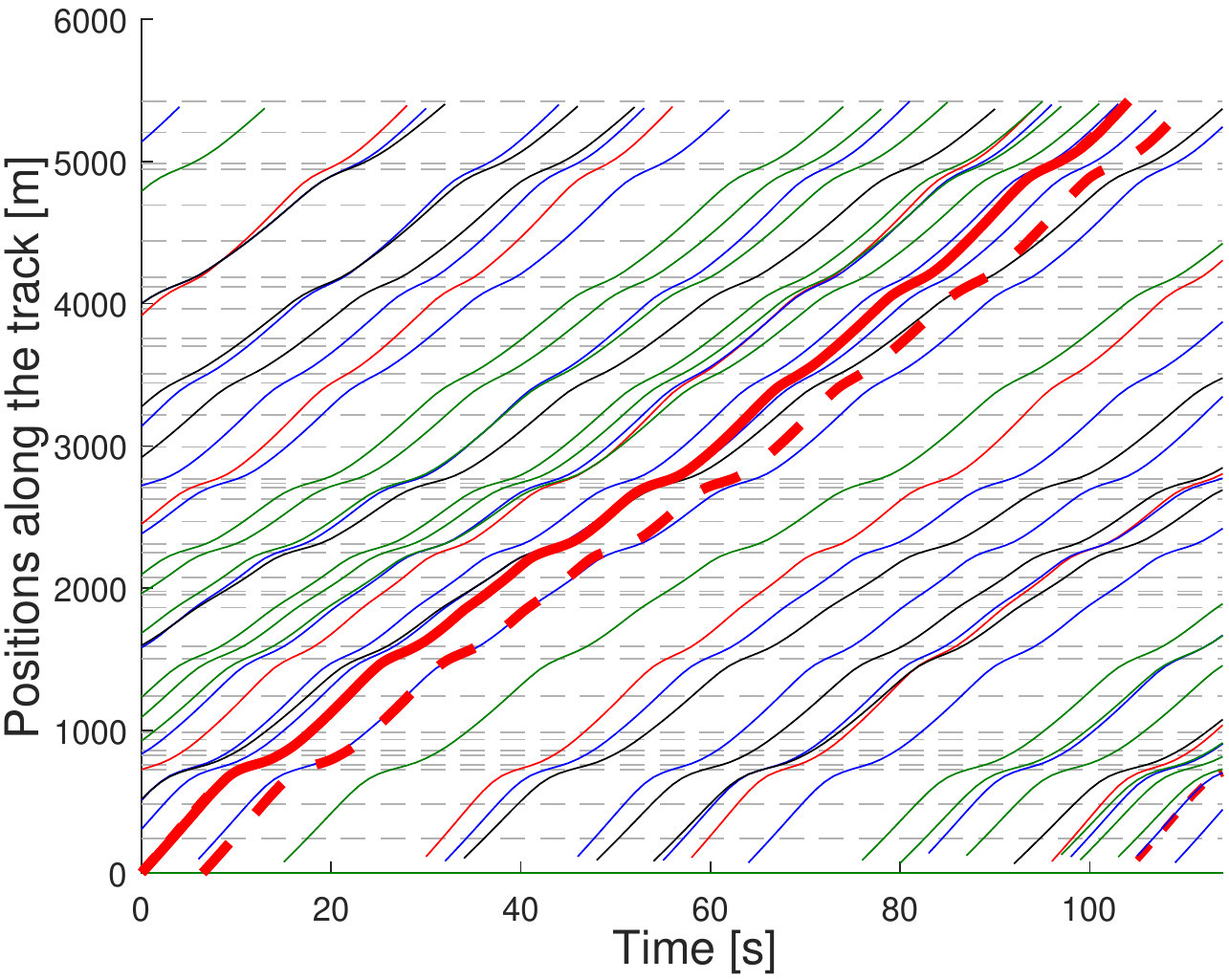}}
	\caption{Result from the application of Stochastic Dynamic Programming to an entire stint of the race. The red thick line represents the space profile deriving from the suggested strategy at the final lap of the stint, whereas the red thick dotted line represents the positions occupied by the reference car during the last lap of the stint in the real race. The other lines represent the competitors' motion. Adopting our approach to an entire stint of 26 laps, a time gain of 6.44 s would have been achieved.}
	\label{fig:stint_result}
\end{figure}

\begin{figure}[htb]
	\centerline{\includegraphics[width=0.95\columnwidth]{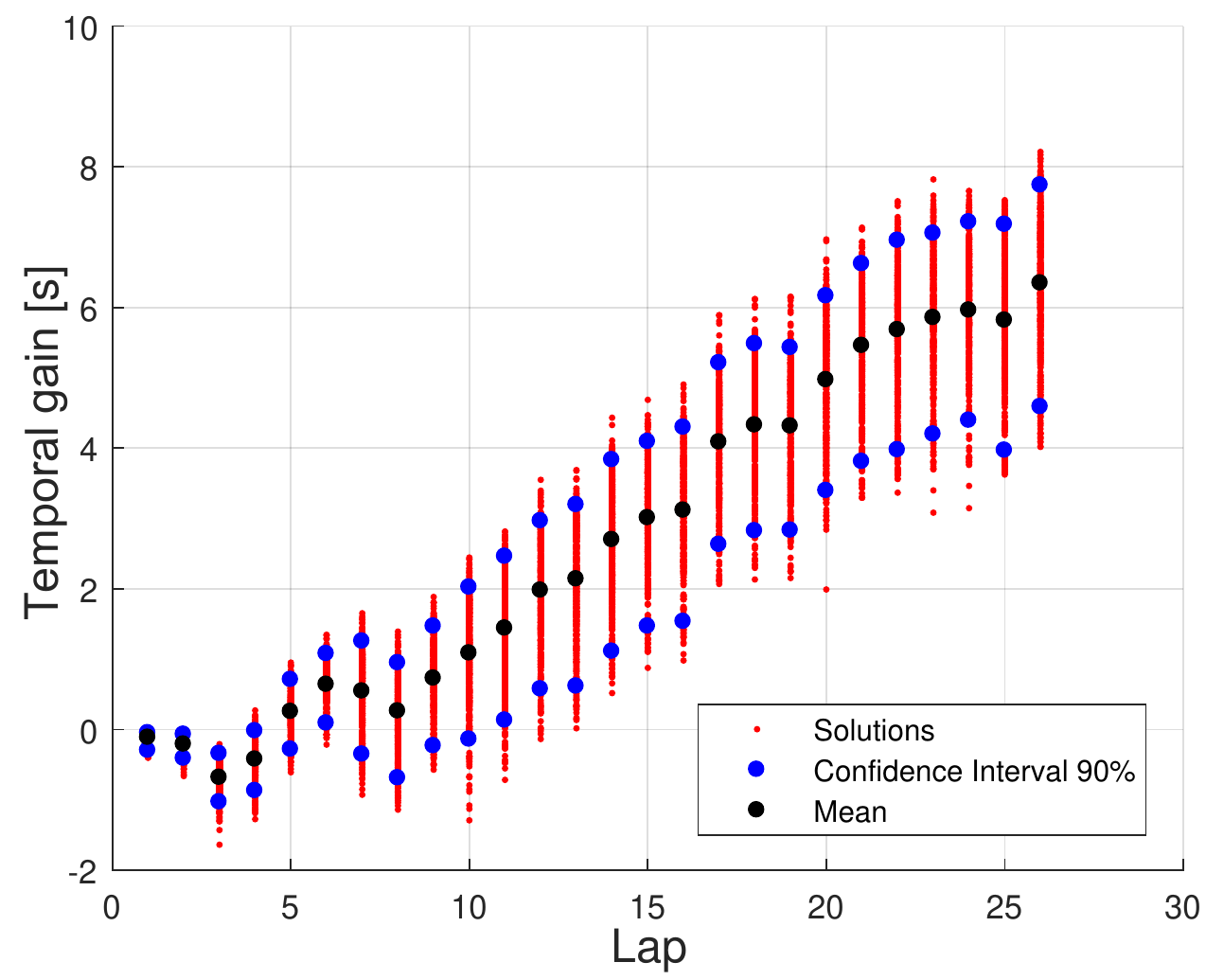}}
	\caption{Lap-by-lap temporal gain generated by our approach.}
	\label{fig:stint_temp_gain}
\end{figure}

The analysis can be extended to a sequence of laps, i.e. a stint of the race. Considering the fifth stint of the Bahrain 2017 event, we launch a Monte Carlo numerical simulation starting with the initial positions of the competitors, and identify the best among the fifteen pre-computed strategies through SDP. The positions of the ego-car during the real race are then replaced by the space profile deriving from the optimal strategy. Then, SDP is employed a second time to evaluate the statistical time loss due to overtakings that would have been generated through the optimal strategy. Figure \ref{fig:stint_result} shows that our method can lead to a time gain of $6.44$ s compared to the original strategy used during the race.  
Running the procedure multiple times, we obtained an upper bound equal to $6.44 + 1.44$ s and a lower bound of $6.44-1.83$ s, with confidence interval $90\%$. Figure \ref{fig:stint_temp_gain} represents the temporal gain lap by lap, with associated confidence interval. It is evident that the application of the proposed method can have a significant impact on the result of an event, since the time difference between the winner and the competitors is usually of a few seconds. 

\section{Conclusions}
\label{sec: conclusions}
In this work, we have presented an efficient procedure to generate optimal lap strategies for LMP1 hybrid electric vehicles in WEC events. Particularly, the framework computes the optimal energy budget utilisation that statistically minimises the ego-vehicle lap time in presence of competitors, while complying with the technical regulations. 

Our approach relies on multiple contributions. First, we have shown how to extract meaningful statistics regarding the competitors' motion along the track, using the limited amount of publicly available data from previous races. The statistics model the sector times and overtaking probability distributions of each vehicle type. They are used to develop realistic Monte Carlo simulations of the agents motion and interactions in a lap. Second, we have computed a set of candidate traffic-free solutions to the ego-vehicle lap strategy optimisation problem using Genetic Algorithms, presenting them as a more computationally efficient alternative to the classical MIQCP formulation. Finally, the traffic-aware optimal solution is statistically identified among the candidate traffic-free policies using Stochastic Dynamic Programming. Relying on several Monte Carlo simulations of the competitors' motion, Stochastic Dynamic Programming is used to evaluate the best strategy among the candidate ones.

To validate our approach, we have applied it to a stint of a real race. The results show that our approach leads to an average time gain of $6.44$ s, compared to the strategy that was actuated during the real race. The time gain is particularly significant, since the time difference between the winner and the leading vehicles is usually of a few seconds in WEC events.

Finally, we highlight that our strategy can be easily extended to other types of racing events. In fact, it only relies on an ego-vehicle longitudinal model and on the sector times data of the vehicles from previous races, which are commonly available. The extension to different types of events is left to future works.

\appendices

\section*{Acknowledgment}
This work has been developed in collaboration with Dr. Ing. h.c. F. Porsche AG. We would like to thank the Porsche Motorsport Team for providing sensitive data on WEC events and their vehicle. 
A sincere thanks to Stephen Mitas, Porsche Chief Race Engineer, and to Emiliano Giangiulio and Vincenzo Scali, Porsche engineers, for 
contributing to the establishment of this work and for their support. 

\bibliographystyle{IEEEtran}
\bibliography{IEEEabrv,main}

\end{document}